\documentclass[11pt,twoside]{article}

\usepackage{graphicx}
\usepackage{fancyhdr}
\usepackage{amsmath}
\usepackage{amssymb}
\usepackage{color}

\usepackage[textwidth=16cm,textheight=22cm,top=3cm,bottom=3cm]{geometry}

\makeatletter
\def\BState{\State\hskip-\ALG@thistlm}
\makeatother

\newtheorem{theorem}{Theorem}
\newtheorem{remark}{Remark}
\newtheorem{corollary}{Corollary}
\newtheorem{lemma}{Lemma}

\graphicspath{{./figuras_arxiv/}}

\newcommand{\drop}[1]{}
\newcommand{\no}{\noindent}
\newcommand{\fer}[1]{(\ref{#1})}
\newcommand{\qtext}[1]{\quad\text{#1}}

\newcommand{\bc}{\mathbf{c}}

\newcommand{\bx}{\mathbf{x}}
\newcommand{\by}{\mathbf{y}}

\newcommand{\cK}{\mathcal{K}}

\newcommand{\eps}{\varepsilon}
\newcommand{\vfi}{\varphi}

\newcommand{\grad}{\nabla}

\newcommand{\p}{\partial}

\newcommand{\N}{\mathbb{N}}
\newcommand{\R}{\mathbb{R}}

\def\O{\Omega}
\newcommand{\X}{{\cal X}}

\newcommand{\Chi}[1]{\chi_{\scriptscriptstyle #1}}

\newcommand{\abs}[1]{| #1 |}
\newcommand{\nor}[1]{\| #1 \|}

\renewcommand{\P}{P$(\O,u_0)$}
\newcommand{\Ps}{P$(\O_*,u_{0*})$}

\DeclareMathOperator{\NF}{NF}
\DeclareMathOperator{\sign}{sign}
\begin{document}
%
%
%

\title{Well-posedness of a nonlinear integro-differential problem and its rearranged formulation 
\thanks{To appear in Nolinear Analysis Real World Applications (2016)}
 \thanks{First and third authors supported by the Spanish MCI Project MTM2013-43671-P. 
 Second author supported by the Spanish Project TEC2012-39095-C03-02}}

\author{Gonzalo Galiano  \thanks{Dpt. of Mathematics, Universidad de Oviedo,
 c/ Calvo Sotelo, 33007-Oviedo, Spain ({\tt galiano@uniovi.es, julian@uniovi.es})}
 \and Emanuele Schiavi \thanks{Dpt. of Mathematics, Universidad Rey Juan Carlos,
  Madrid, Spain ({\tt emanuele.schiavi@urjc.es})}
    \and Juli\'an Velasco\footnotemark[2] }
\date{}

\pagestyle{fancy}
\fancyhead{}
\fancyhead[LE]{G. Galiano, E. Schiavi, and J. Velasco} 
\fancyhead[RO]{Well-posedness of a nonlinear integro-differential problem and its rearranged formulation}
\thispagestyle{plain}

\maketitle

\begin{abstract}
We study the existence and uniqueness of solutions of a nonlinear integro-differential problem
which we reformulate introducing the notion of the decreasing 
rearrangement of the solution. A dimensional reduction of the problem is obtained and a detailed analysis of the properties of the solutions of the model is provided.
Finally, a fast numerical method is devised and implemented to show the performance of the model when typical image processing tasks such as 
filtering and segmentation are performed.

\no\emph{Keywords: }{Integro-differential equation, existence, uniqueness, neighborhood filters, decreasing rearrangement, denoising, segmentation.}
\end{abstract}

\section{Introduction}\label{intro}
This article is devoted to the study of the nonlinear integro-differential problem 
\begin{align}
\label{eq.orig}
  \p_t u(t,\bx)  = & \int_\O \cK_h (u(t,\by)-u(t,\bx))(u(t,\by)-u(t,\bx)) d\by \\
   &+\lambda(u_0(\bx)-u(t,\bx)), \nonumber \\
    u(0,\bx)= & u_0(\bx) \label{eq.orig2}
 \end{align}
 for $(t,\bx)\in Q_T=(0,T)\times\O$. Here, $\O\subset\R^d$ $(d\geq 1)$ denotes an open and bounded set, $T>0$, $\lambda >0$ and  $u_0\in BV(\O)\cap L^\infty(\O)$. 
  The {\em range} kernel $\cK_h$ is given as a rescaling 
 $\cK_h(\xi)= \cK(\xi/h)$ of a kernel $\cK$ satisfying the usual properties of nonnegativity and smoothness. We shall give the precise assumptions in Section~\ref{sec.results}.
  We shall refer to problem \fer{eq.orig}-\fer{eq.orig2} as to problem P$(\O,u_0)$.
  The main results contained in this article are:
 \begin{itemize}
  \item Theorem~\ref{th.existenceU}. The well-posedness of problem P$(\O,u_0)$, the stability property of its solutions with respect to the initial datum, and the time invariance of the level set structure of its solutions.
  
  \item Theorem~\ref{th.equiv}. The equivalence between solutions of problem P$(\O,u_0)$ and the one-dimensional problem \Ps, where $\O_*=(0,\abs{\O})$, and $u_{0_*}$ is the decreasing rearrangement of $u_0$, see Section~\ref{sec.dec_re} for definitions.
  
  \item Theorem~\ref{th.pde}. The asymptotic behavior of the solution of problem P$(\O_*,u_{0*})$ with respect to 
  the window size parameter, $h$, as a shock filter. 
 \end{itemize}

 Problem \P~is related to some problems arising in Image Analysis, Population Dynamics and other disciplines. The general formulation in (\ref{eq.orig}) includes, for example, a time-continuous 
 version of the Neighborhood filter (NF) operator:
\begin{equation*}
 \NF^h u (\bx)=\frac{1}{C(\bx)}\int_\O \textrm{e}^{-\frac{\abs{u(\bx)-u(\by)}^2 }{h^2}} u(\by)d\by,
\end{equation*}
where $h$ is a positive constant, and $C(\bx)=\int_\O \exp\left(-\abs{u(\bx)-u(\by)}^2) h^{-2}\right) d\by$ is a normalization factor. In terms of the notation introduced for problem \P\, 
 the NF is recovered setting $\cK(s)= \exp(-s^2)$ and $\lambda =0$.
This  well known denoising filter is usually employed in the image community through an  iterative scheme,
\begin{equation}
\label{def.GFV}
 u^{(n+1)} (\bx)=\frac{1}{C_n(\bx)}\int_\O \cK_h(u^{(n)}(\bx)-u^{(n)}(\by)) u^{(n)}(\by)d\by,
\end{equation}
 with $C_n(\bx)=\int_\O \cK_h(u^{(n)}(\bx)-u^{(n)}(\by)) d\by$. It is the simplest particular case of other related filters involving nonlocal terms, 
notably the Yaroslavsky filter \cite{Yaroslavsky1985,Yaroslavsky2003}, the Bilateral filter  \cite{Smith1997,Tomasi1998}, and the Nonlocal Means filter \cite{Buades2005}.

These methods have been introduced in the last decades
as efficient alternatives to local methods such as those expressed in terms 
of nonlinear diffusion partial differential equations (PDE's), among which the pioneering nonlinear anti-diffusive model of Perona and Malik \cite{Perona1990}, 
the theoretical approach of \'Alvarez et al. \cite{Alvarez1992} and the celebrated ROF model of Rudin et al. \cite{Rudin1992}. 
We refer the reader to \cite{Buades2010} for a review comparing these local and non-local methods.

Another image processing task encapsulated by problem \P~is the {\em histogram prescription}, used for image contrast enhancement: 
Given an initial image $u_0$, find a companion image $u$ such that $u$ and $u_0$ share the same level sets structure, and the histogram distribution of $u$ is given by a prescribed function $\Psi$. A widely used choice is $\Psi(s)=s$, implying that $u$ has a uniform histogram distribution. In this case, $\cK(s)= sign^{-}(s)/s$ 
and $\lambda$ is related to the image size and its dynamic range, 
see Sapiro and Caselles \cite{Sapiro1997} for the formulation and analysis of the problem. 
Nonlinear integro-differential  of the form 
\begin{align}
\label{mazon}
 \p_t u(t,\bx)  =\int_\O  (u(t,\by)-u(t,\bx)) w(\bx-\by)d\by 
\end{align}
and other nonlinear variations of it have also been recently used (Andreu et al. \cite{Andreu2010}) to model diffusion processes in Population Dynamics and other areas. More precisely,  if $u(t,\bx)$ is thought of as a density at the point
$\bx$ at time $t$ and $w(\bx-\by)$ is thought of as the probability distribution of jumping from location $\by$ to location $\bx$, then $\int_\O  u(t,\by) w(\bx-\by)d\by $  is the rate at which individuals are arriving at position $\bx$ from all other places and $-u(t,\bx) = - \int_\O  u(t,\bx) w(\bx-\by)d\by$  is the rate at which they are leaving location $\bx$. 
In the absence of external or internal sources this consideration leads immediately to the fact that the density $u$ satisfies the equation \fer{mazon}. 

These kind of equations are called nonlocal diffusion equations since in them the
diffusion of the density $u$ at a point $\bx$ and time $t$ depends not only on $u(t, \bx)$ but also on the values of $u$ in a set determined (and weighted) by the space kernel $w$. A thoroughfull study of this problem may be found in the monograph by Andreu et al. \cite{Andreu2010}. 
Observe that in problem \P, the space kernel is taken as $w\equiv 1$, meaning that the influence of nonlocal diffusion is spread to the whole domain. 

As noticed by Sapiro and Caselles \cite{Sapiro1997} for the histogram prescription problem, and later by  Kindermann et al. \cite{Kindermann2005} for the iterative Neighborhood filter  \fer{def.GFV}, or by Andreu et al. \cite{Andreu2010} for continuous time problems like \fer{mazon},  these formulations may be deduced from variational considerations. For instance, in \cite{Kindermann2005}, the authors consider, for $u\in L^2(\O)$, the functional
\begin{equation}
\label{def.functional}
 J(u)=\int_{\O\times\O} g (u(\bx)-u(\by)) w(\bx-\by) d\bx d\by,
\end{equation}
with an appropriate spatial kernel $w$, and a differentiable filter function $g$. Then, the authors formally deduce the equation for the critical points of $J$. 
These critical points  coincide with the fixed points of the nonlocal filters they study. For instance, if $g(s)=\int_0^s\cK_h (\sqrt{\sigma}) d\sigma $ and $w\equiv 1$, the critical points satisfy
\begin{equation*}
 u(\bx)= \frac{1}{C(\bx)}\int_{\O}\cK_h(u(\bx)-u(\by)) u(\by)d\by,
\end{equation*}
which can be solved through a fixed point iteration mimicking the iterative Neighborhood filter  scheme \fer{def.GFV}. On the other hand, choosing $g(s)=s$ (or some suitable nonlinear variant) and considering a gradient descent method to approximate the stationary solution, equation \fer{mazon}
 is deduced. Similarly, $g(s)=\abs{s}$ and $w\equiv 1$ leads to the histogram prescription problem.

Although the functional \fer{def.functional} is not convex in general, 
Kindermann et al. prove that when $\cK$ is the Gaussian kernel then 
the addition to $J$ of a convex fidelity term, e.g.
\begin{equation*}
 \tilde J (u;u_0)=J(u)+\lambda\nor{u-u_0}^2_{L^2(\O)},
\end{equation*}
gives, for $\lambda>0$ large enough, a convex functional  $\tilde J$, see \cite[Theorem 3.1]{Kindermann2005}. 

Thus, the functional $\tilde J$ may be seen as the starting point for the deduction of problem \P, representing the continuous gradient descent formulation of the minimization problem
modeling Gaussian image denoising. 
Notice that although the convexity of $\tilde J$ is only ensured for $\lambda$ large enough, the results obtained in this article are independent of such value, 
and only the usual non-negativity condition on $\lambda$ is assumed.

The outline of the article is as follows. In Section~\ref{sec.dec_re}, we introduce some basic notation and the definition of \emph{decreasing rearrangement} of a function. This is later used to show the equivalence between the general problem \P~and the reformulation \Ps~ in terms of a problem with a identical structure but defined in a one-dimensional space domain. This technique was already used in \cite{Galiano2015} for dealing with the time-discrete version of problem \P, in the form of the iterative scheme \fer{def.GFV}. See also \cite{Galiano2015b,Galiano2015c} for the problem with non-uniform spatial kernel. 
In Section~\ref{sec.results}, we state our main results. Then, in Section~\ref{sec.numerics}, 
 we introduce a discretization scheme for the efficient approximation of solutions of problem \P, and demonstrate its performance with some examples.
 In Section~\ref{sec.proofs}, we provide the proofs of our results, and finally, in 
 Section~\ref{sec.conclusions}, we give our conclusions.

\section{The decreasing rearrangement}\label{sec.dec_re}
Given an open and bounded (measurable) set $\O\subset\R^d$, $(d\geq 1)$ 
let us denote by $\abs{\Omega}$ its Lebesgue measure and set 
$\O_*=(0,\abs{\O})$.
For a Lebesgue measurable function $u:\O\to\R$, the function 
$q\in\R\to m_u(q) = \abs{\{\bx \in\O : u(\bx) >q\}}$ is called the \emph{distribution function} corresponding to $u$. 
Function $m_u$ is, by definition, non-increasing and therefore admits a unique  generalized inverse, called its
\emph{decreasing rearrangement}. This inverse takes the usual pointwise meaning when 
the function $u$ has not flat regions, i.e. when $\abs{\{\bx \in\O : u(\bx) =q\}} =0$ for any $q\in\R$. In general, 
the decreasing rearrangement $u_*:\bar\O_*\to\R$ is given by:
\begin{equation*}
u_*(s) =\left\{
\begin{array}{ll}
 {\rm ess}\sup \{u(\bx): \bx \in \O \} & \qtext{if }s=0,\\
 \inf \{q \in \R : m_u(q) \leq s \}& \qtext{if } s\in \O_*,\\
 {\rm ess}\inf \{u(\bx): \bx \in \O \} & \qtext{if }s=\abs{\O}.
\end{array}\right.
 \end{equation*}
Notice that since $u_*$ is non-increasing in $\bar\O_*$, it is continuous but at most a countable subset of  
$\bar\O_*$. In particular, it is right-continuous for all $\sigma\in (0,\abs{\O}]$.

The notion of rearrangement of a function is classical and was introduced by Hardy, Littlewood and Polya \cite{Hardy1964}. Applications include the study of isoperimetric and variational inequalities \cite{Polya1951,Bandle1980,Mossino1984,Mossino1986}, comparison of solutions of partial differential equations \cite{Talenti1976,Alvino1978,Vazquez1982,Diaz1985,Diaz1995,Alvino1996}, and others.
We refer the reader to the monograph \cite{Rakotoson2008} for a extensive research on this topic.

Two of the most remarkable properties of the decreasing rearrangement are the  equi-measurability property 
\begin{equation}
\label{prop.1}
 \int_\O f(u(\by))d\by = \int_0^{\abs{\O}} f(u_* (s))ds,
\end{equation}
for any Borel function $f:\R\to\R_+$, and the contractivity
\begin{equation}
 \label{prop.2}
 \nor{u_*-v_*}_{L^p(\O_*)}\leq \nor{u-v}_{L^p(\O)},
\end{equation}
for $u,v\in L^p(\O)$, $p\in[1,\infty]$.

For the extension of the decreasing rearrangement to families of functions depending on a parameter, e.g. $t\in [0,T]$, we first consider, 
for $t$ fixed, the function $u(t):\O\to\R$ given by $u(t)(\bx)=u(t,\bx)$, for any $\bx\in\O$. Then we define $u_*:(0,T)\times\O_*\to\R$ by
$ u_*(t,s)=u(t)_*(s)$.

\section{Main results}\label{sec.results}

Our first result ensures the well-posedness of problem \P~ for $L^\infty(\O)$ initial data with bounded total variation. In addition, we show that the level sets structure of the solution is time invariant. 
Before stating our results, we collect here the main assumptions on the data problem, to which we shall refer to as \textbf{(H)}:
\begin{itemize}
 \item $\O\subset\R^d$ is an open, bounded, and connected set ($d\geq 1$). 
\item  The final time, $T$, which simulate the time horizon of the diffusion process is a real, positive fixed number.
\item  The parameter $\lambda$  is a  real, nonnegative fixed number.
 
 \item $\cK \in W^{1,\infty}(\R)$ is nonnegative.

 \item $u_0\in \X:=L^\infty(\O)\cap BV(\O)$ is assumed to be, without loss of generality, non-negative.
\end{itemize}
Basic facts but also advanced results about the space of bounded variation $BV (\Omega )$ can be found in the book by Ambrosio et al. \cite{Ambrosio2000}.
Notice that, depending on the space dimension $d \geq 2$ we have the continuous injections $BV (\Omega )\hookrightarrow L^{d/d-1}(\Omega )$.  
When $d=1$ we have $\X \equiv BV(\O)$.

\begin{theorem}
 \label{th.existenceU}
Assume (H).  Then there exists a unique solution 
$u\in C^\infty([0,T];\X)$ of problem \P. 
In addition, if $u_{01},u_{02}\in \X$ and 
$u_1,u_2 \in C^\infty([0,T];\X)$ are the corresponding solutions to problems 
$P(\O,u_{01}),~P(\O,u_{02})$ then
\begin{equation}
 \label{stability}
 \nor{u_1-u_2}_{L^\infty(0,T;L^2(\O))}\leq C\nor{u_{01}-u_{02}}_{L^2(\O)},
\end{equation}
for some constant $C>0$.

Finally, suppose that $u_0(\bx_1)=u_0(\bx_2)$ for some  $\bx_1,\bx_2\in\O$.
Then $u(t,\bx_1)=u(t,\bx_2)$ for all $t\in (0,T]$. 
\end{theorem}

\begin{remark}
The existence and stability results of Theorem~\ref{th.existenceU} may be extended to more general zero-order terms in the equation 
\fer{eq.orig} of problem \P. For instance, we can consider a function $f:[0,T]\times\O\times\R\to\R$ satisfying 
$f(\cdot,\bx,s)\in L^\infty(0,T)$, $f(t,\cdot,s)\in BV(\O)$, and $f(t,\bx,\cdot)\in W^{1,\infty}(\R)$. This regularity coincides with the initially obtained for the integral term of equation \fer{eq.orig} in the approximation procedure to construct the solution.
In addition, if $u_0(\bx_1)=u_0(\bx_2)$ implies $f(\cdot,\bx_1,\cdot)=f(\cdot,\bx_2,\cdot)$, then the time invariance of level sets holds.
\end{remark}

 Replacing the set $\O$ by $\O_*$ and the initial data $u_0\in \X$  
 by $v_0\in \X_*\equiv BV(\O_*)$, Theorem~\ref{th.existenceU}  ensures the existence 
 of a solution of problem \Ps. 
 Observe that  $\O_*\subset\R$ is bounded because  $\O\subset\R^d$ is bounded (assumption $(H)$)  and this implies  $BV(\O_*) \subset L^\infty(\O_*)$ and $\X_* \equiv  BV(\O_*)$.
  
In the following result we obtain some properties of solutions of the one-dimensional problem P$(\O_*,v_0)$. Although the corollary is valid for any interval in $\R$, we keep the notation 
$\O_*$ for simplicity. The corresponding result for the discrete-time version, with $\lambda=0$, of problem \Ps~may be found in \cite{Galiano2015}.

\begin{corollary}
\label{th.existenceV}
Assume (H), and let $v\in C^\infty ([0,T];\X_*)$ be the solution of 
problem P$(\O_*,v_0)$, for some nonincreasing $v_0 \in\X_*$. Then 
\begin{enumerate}
 \item $\sign(\p_s v(t,\cdot))=\sign(\p_s v_0)\leq 0$ 
a.e. in $\O_*$, for all $t\in(0,T)$. 

\item For $t>0$, $v(t,0)\leq v_0(0)$ and $v(t,\abs{\O})\geq v_0(\abs{\O})$.

\item If $\cK$ is odd then  $\int_{\O_*} v(t,s)ds =\int_{\O_*} v_0(s)ds$, for $t>0$.

\item If $\cK\in W^{m,\infty}(\R)$ and $v_0\in W^{m,p}(\O_*)$, for $m\in\N$ and $1<p<\infty$, then $v\in C^\infty([0,T];W^{m,p}(\O_*))$.

\item If $\lambda=0$ and $\cK \in C^1(\R)$ is such that $\cK'_h(\xi)\xi + \cK_h(\xi) >0$ then 
$v(t,\cdot)\to const.$ as $t\to\infty$.

\end{enumerate}

\end{corollary}
\begin{remark}
 Condition in point 3 is a natural symmetry condition for convolution kernels and it is satisfied, for instance, by the Gaussian kernel. Condition in point 5 is also satisfied by the Gaussian kernel, if $h$ is large enough.
\end{remark}

The next result establishes the connection between problems \P~and \Ps.

\begin{theorem}
\label{th.equiv}
Assume (H).
Then, $u \in C^\infty([0,T];\X)$ is a solution of \P~if and only if $u_* \in C^\infty([0,T];\X_*)$ is a  solution of \Ps. 
\end{theorem}

Theorem~\ref{th.equiv} implies that the solution of the multi-dimensional problem 
\P~ may be constructed by solving the one-dimensional problem \Ps. Indeed, using 
the level sets invariance asserted in Theorem~\ref{th.existenceU}, we deduce
\begin{equation*}
 u(t,\bx)=u_*(t,s)\qtext{for a.e. }\bx\in \{ \by\in\O: u_0(\by)=u_{0*}(s)\},
\end{equation*}
for all $t\in [0,T]$. When image processing applications are considered, by property 1 of Corollary~\ref{th.existenceV}, the solution to \P~may be understood as a \emph{contrast change} of the initial image, $u_0$.

Indeed, this property also implies that if, initially, $u_0$ has no flat regions, 
and therefore $u_{0*}$ is decreasing, then the solution of \Ps~verifies this property for all $t>0$. Then, Theorem~\ref{th.existenceV} implies that the solution of \P~has no flat regions for all $t>0$.

The last theorem is an extension of a result given in \cite{Galiano2015} for the discrete-time formulation with $\lambda=0$. In it, we deduce the asymptotic behavior of the solution $u_*$ of problem \Ps~(and thus of $u$ of problem \P) in terms of the window size parameter, $h$. 
Although we state it for the Gaussian kernel, more general choices are possible, see \cite[Remark 2]{Galiano2015}.

\begin{theorem}
\label{th.pde}
Assume (H) with $ \cK(\xi)=\text{e}^{-\xi^2}$ and $u_0\in \X$ having no flat regions.
Suppose, in addition, that $u_{0*} \in  C^3(\bar\O_*)$.
Then, for all $(t,s) \in [0,T]\times \O_*$, there exist positive constants $\alpha_1,\alpha_2$ independent of $h$ such that the solution $u_*\in C^\infty([0,T];C^3(\bar\O_*))$ of \Ps~satisfies
\begin{equation}
\label{app.NFstar}
\p_t u_*(t,s)= \lambda(u_{0*}(s)-u_*(t,s))+ \alpha_1 \tilde k_h(t,s) h^2-
  \alpha_2 \frac{\p^2_{ss} u_*(t,s) }{\abs{\p_s u_*(t,s)}^3} h^3 + O(h^{7/2}),
\end{equation}
with 
\begin{equation}
\label{def.ktilde}
 \tilde k_h(t,s)= \frac{\cK_h(u_*(t,\abs{\O})-u_*(t,s))}{\abs{\p_s u_*(t,\abs{\O})}}-
 \frac{\cK_h(u_*(t,0)-u_*(t,s))}{\abs{\p_s u_*(t,0)}},
\end{equation}
and with $\alpha_1\approx 1/(2\sqrt{\pi})$, and $\alpha_2\approx 1$.
\end{theorem}

Two interesting effects captured by \fer{app.NFstar} are the following:
 \begin{enumerate}
  \item The border effect (range shrinking). Function $\tilde k_h$ is \emph{active} only when $s$ is close to the boundaries, $s\approx 0$ and $s\approx \abs{\O}$. For 
   $s \approx 0$, $\tilde k_h(t,s) <0 $ contributes to the decrease of 
  the largest values of $u_*$ while for $s \approx \abs{\O}$ we have $\tilde k_h(t,s) >0$,
  increasing the smallest values of   $u_*$. Therefore, 
  this term tends to flatten $u_*$. In image processing terms, a loss of contrast is induced.
  
  \item The term 
  \[
-\frac{\p^2_{ss} u_*(t,s) }{\abs{\p_s u_*(t,s)}^3}   
  \]
 is anti-diffusive, inducing large gradients on $u_*(t,\cdot)$ in a neighborhood of  inflexion points. In this sense, the scheme \fer{app.NFstar}
  is related to the shock filter introduced by \'Alvarez and Mazorra \cite{Alvarez1994} 
  \begin{equation}
  \label{alvarez}
   v_t+F(G_\sigma v_{xx},G_\sigma v_{x})v_x=0,
  \end{equation}
where $G_\sigma$ is a smoothing kernel and function $F$ satisfies $F(p,q)pq\geq 0$ for any $p,q\in\R$. Indeed, neglecting the fidelity, the border and the lower order terms, and defining $F(p,q)=\frac{p}{q\abs{q}^3}$, we render \fer{app.NFstar} to the form
\fer{alvarez}.

This property can be exploited to produce a partition of the image so the model can be interpreted as a tool for fast segmentation and classification.  
An example is proposed in the numerical experiments where a time-continuous version of the NF is implemented.
 \end{enumerate}

 \section{Discretization and numerical examples}\label{sec.numerics}

For the discretization of problem \P, for $u_0:\O\subset\R^d\to\R$, we take advantage of the equivalence result stated in Theorem~\ref{th.equiv}. Thus, we first calculate a numerical approximation, $\tilde u_{0*}$, to the decreasing rearrangement $u_{0*}:\O_*\subset\R\to\R$ and consider the problem P$(\O_*,\tilde u_{0*})$. Then, we discretize this one-dimensional problem and compute a numerical approximation, $v:[0,T]\times\O_*\to\R$. By   Theorem~\ref{th.equiv}, $v$ is, in fact, an approximation to $u_*$, where $u:[0,T]\times\O\to\R$ is a solution to problem \P. Then, we finally recover an approximation, $\tilde u$, to $u$ by defining 
\begin{equation}
\label{def.u5}
 \tilde u(t,\bx)=v(t,s)\qtext{for a.e. }\bx\in \{ \by\in\O: \tilde u_0(\by)=\tilde u_{0*}(s)\}.
\end{equation}

Inspired by the image processing application of problem \P, we consider a piecewise constant approximation 
to its solutions. 
Let $\O\subset \R^2$ be, for simplicity, a rectangle domain and consider a uniform 
mesh on $\O$ enclosing square elements (pixels), $T_{mn}$, of unit area, with barycenters  denoted by $\bx_{mn}$, for $m=1,\ldots,M$ and $n=1,\ldots,N$. Given $u_0\in L^\infty(\O)\cap BV(\O)$, we consider its piecewise constant interpolator $\tilde u_0(\bx) = u_0(\bx_{mn})$ if $\bx\in T_{mn}$. 

The interpolator $\tilde u_0$ has a finite number, $Q\in\N$, of quantized levels that we denote by $q_i$, 
with $\max (\tilde u_0)=q_1 >\ldots > q_Q=\min (\tilde u_0)$. That is $\tilde u_0(\bx)=\sum_{j=1}^Q q_j\chi_{E_j} (\bx),$
where $E_j$ are the level sets of $\tilde u_0$, 
 \begin{equation*}
E_j=\{\bx \in \O : \tilde u_0(\bx)=q_j \},\quad j=1,\ldots,Q.
\end{equation*}
Since $\tilde u_0$ is piecewise constant, the decreasing rearrangement of $\tilde u_0$ is 
piecewise constant too, and given by 
\begin{equation}
\label{def.u02}
 \tilde u_{0*}(s)=\sum_{j=1}^Q q_j\chi_{I_j}(s), 
\end{equation}
with $I_j=[a_{j-1},a_j)$ for $j=1,\ldots,Q$, and $a_0=0$, $a_1=\abs{E_1}$, $a_2=\abs{E_1}+\abs{E_2}$,$\ldots$,$a_Q=\sum_{j=1}^Q \abs{E_j}=\abs{\O}$. 

Let $v$ be a candidate to solve problem P$(\O_*,\tilde u_{0*})$.  
Due to the time-invariance of the level sets structure of the solution to this problem, see Theorem~\ref{th.existenceU}, we may express $v$ as
\begin{equation}
\label{sol.v}
 v(t,s)=\sum_{j=1}^Q c_j(t)\chi_{I_j}(s), 
\end{equation}
with $c_1(t)\geq \ldots \geq c_Q(t)$, for $t\in (0,T]$, $c_j(0)\equiv c_j^0=q_j$, for $j=1,\ldots,Q$. Substituting $v$ in equation \fer{eq.orig}, we get, for $s\in  I_j$ and $j=1,\ldots,Q$,
\begin{align}
\label{eq.dis}
c_j'(t)=\sum_{k=1}^Q \cK_h(c_k(t)-c_j(t))(c_k(t)-c_j(t)) \mu_k +\lambda (c_j^0-c_j(t)),
\end{align}
with $\mu_k=a_k-a_{k-1}=\abs{E_k}$. 
Since,  by assumptions (H), the right hand side of \fer{eq.dis} is Lipschitz continuous, the existence and 
uniqueness of a smooth $\bc=(c_1,\ldots,c_Q):[0,T]\to \R_+^Q$ satisfying \fer{eq.dis} and $\bc(0)=\bc^0$ follows.

For the time discretization, we take a uniform mesh of the interval $[0,T]$ of size $\tau>0$, and use the notation $\bc^n = \bc(t_n)$, with $t_n=n\tau$, and $n=0,1,2,\ldots$ Then, we consider the following implicit time discretization of problem \fer{eq.dis}. For $j=1,\ldots,Q$ and $n\geq 1$, solve
\begin{align}
\label{eq.dis2}
c_j^{n}=c_j^{n-1}+\tau\sum_{k=1}^Q \cK_h(c_k^{n}-c_j^{n})(c_k^{n}-c_j^{n}) \mu_k +\tau\lambda (c_j^0-c_j^{n}).
 \end{align}
Since problem \fer{eq.dis2} is a nonlinear algebraic system of equations, we use a fixed point argument to approximate its solution, $\bc^{n}$, at each discrete time $t_{n}$, from the previous approximation $\bc^{n-1}$. 
Let $\bc^{n,0}=\bc^{n-1}$. Then, for $m\geq 1$ the problem is to find $\bc^{n,m}$ solving the linear system 
\begin{align}
\label{eq.dis3}
c_j^{n,m}=c_j^{n-1}+\tau\sum_{k=1}^Q \cK_h(c_k^{n,m-1}-c_j^{n,m-1})(c_k^{n,m}-c_j^{n,m}) \mu_k +\tau\lambda (c_j^0-c_j^{n,m}),
\end{align}
for $j=1,\ldots,Q$. We choose the stopping criterion $\nor{\bc^{n,m}-\bc^{n,m-1}}_\infty < {\rm tol}$, for values of ${\rm tol}$ chosen empirically, and then set $\bc^n=\bc^{n,m}$. 

Finally, using formula \fer{def.u5}, the expression of the initial datum \fer{def.u02}, and 
the definition \fer{sol.v},
 we recover a piecewise constant approximation to the original problem, \P, taking 
\begin{equation*}
 \tilde u (t,\bx)=c_j^n \qtext{if } t\in [t_n,t_{n+1}), \quad \bx\in \{ \by\in\O: \tilde u_0(\by)= q_j\}.
\end{equation*}

\subsection{Example. Histogram based image segmentation}
As an application  we consider a Grand Challenge in Biomedical Image Analysis.
This is a computer vision problem in biomedicine which consists of overlapping cells segmentation and subcellular  nucleus and cytoplasm detection, see \cite{lucell}, \cite{cells}.
The dataset was downloaded from the  {\em Overlapping Cervical Cytology Image Segmentation Challenge}\footnote{{\tt http://cs.adelaide.edu.au/$\sim$carneiro/isbi14\_challenge/index.html}}, ISBI 2014.

The data set is composed by  $512\times512$ real and synthetic images containing two or more cells with different degrees of overlapping, contrast, and texture. The phantom images allow the quantitative analysis of segmentation procedures through their  ground-truth, which is carried out by using the Dice similarity coefficient, $DC$: 
for two sets (images) $A$ and $B$, 
\[
 DC = \frac{2\abs{A\cap B}}{\abs{A}+\abs{B}}.
\]
Observe that values of $DC$ close to one indicate high coincidence of the images, that is, of the ground-truth segmentation and the segmentation obtained with our method. 

For running our algorithm, that is, providing an approximation, $\mathbf{c}^n$, of \fer{eq.dis}, 
we consider the usual number of image quantization levels, $Q=256$. The fidelity term is ignored ($\lambda = 0$), and the range window parameter, $h$, is set as $h=25$ for nucleus detection, and as $h=5$ for cytoplasm detection. The tolerance in the fixed point loop \fer{eq.dis3} is taken as ${\rm tol}=1.e-5$. As a stopping criterion, we consider  a combination of a maximum number of time iterations (1000), and an energy stabilization criterion, 
\[
 \abs{J(\mathbf{c}^{n})-J(\mathbf{c}^{n-1})} < 1.e-10,
\]
where $J(\mathbf{c}^n)$ is the discrete version of the functional 
given by \fer{def.functional}, for $w\equiv 1$ and $g(s)=\exp(-s/h)$. Finally, we implement a 
variable time step, $\tau(n)$, inspired by the proof of existence of solutions and given by, for $n\geq 2$, 
\[
 \tau(n) = \frac{\tau(0)}{\abs{J(\mathbf{c}^{n-1})-J(\mathbf{c}^{n-2})}}
\]
with $\tau(1)=\tau(0)=\big(\max_{j}\{\sum_{k=1}^Q \cK_h(c_k^{0}-c_j^{0})\mu_k \}\big)^{-1}$.
In the experiments, we observed that $\tau(n)$ ranges from order $10^{-7}$ in the first iterations to order $10^{-1}$ just before convergence.

We summarize our results for the test90 dataset in Table~\ref{table1}, where we show the $DC$ for some 
specific samples, and the mean $DC$ of the ninety samples contained in the dataset. 
We may check that $DC$ values are very high for the segmentation of both regions of interest (cytoplasm and nucleus), always above the range obtained in \cite{lucell,cells}. The execution times are given for a Matlab implementation of the algorithm, running on a standard laptop (Intel Core i7-2.80 GHz processor, 8GB RAM). 

\begin{figure}[ht] 
\centering 
{\includegraphics[width=3cm,height=2.25cm]{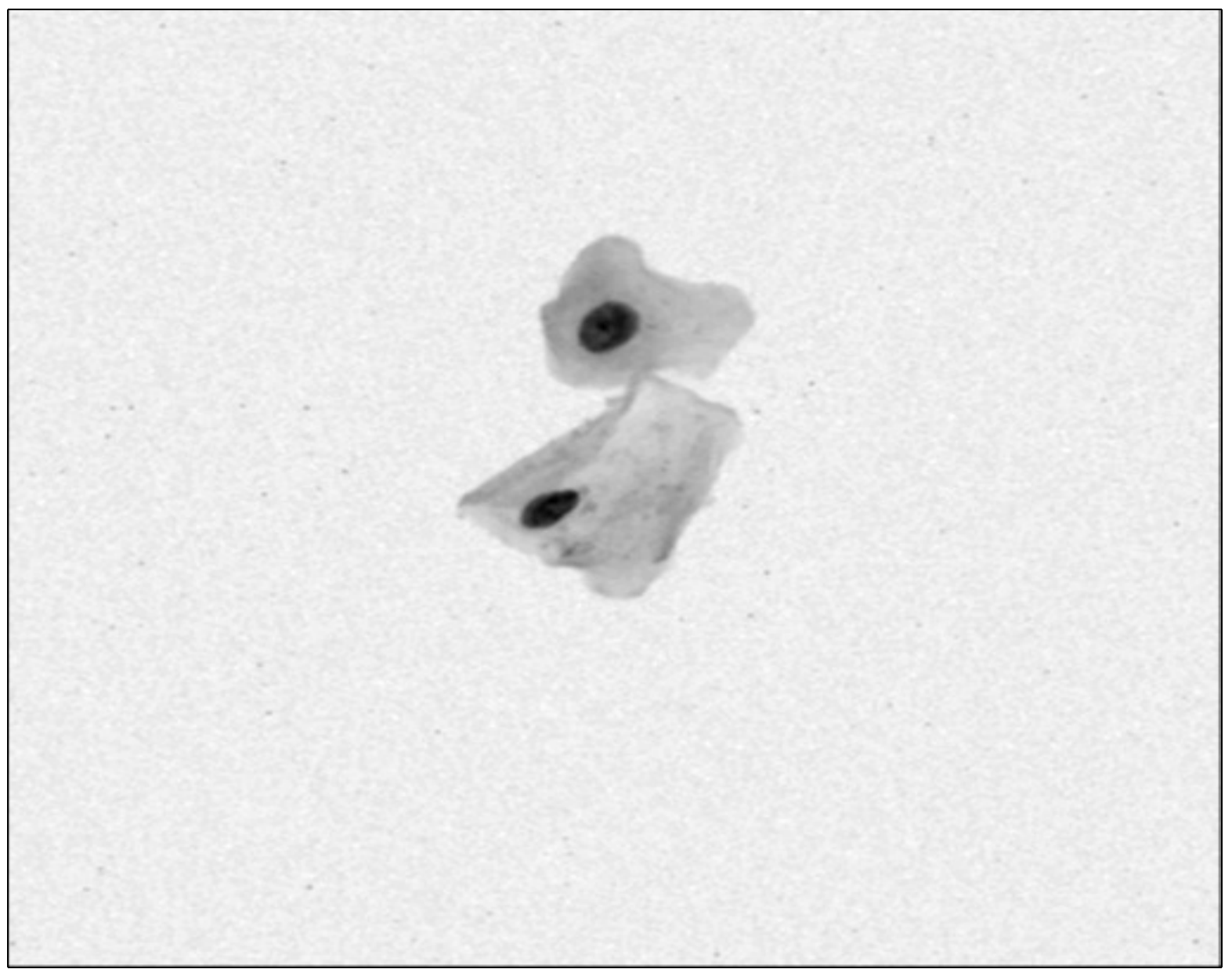}}
{\includegraphics[width=3cm,height=2.25cm]{celulas1_todoSeg-eps-converted-to.pdf}}
{\includegraphics[width=3cm,height=2.25cm]{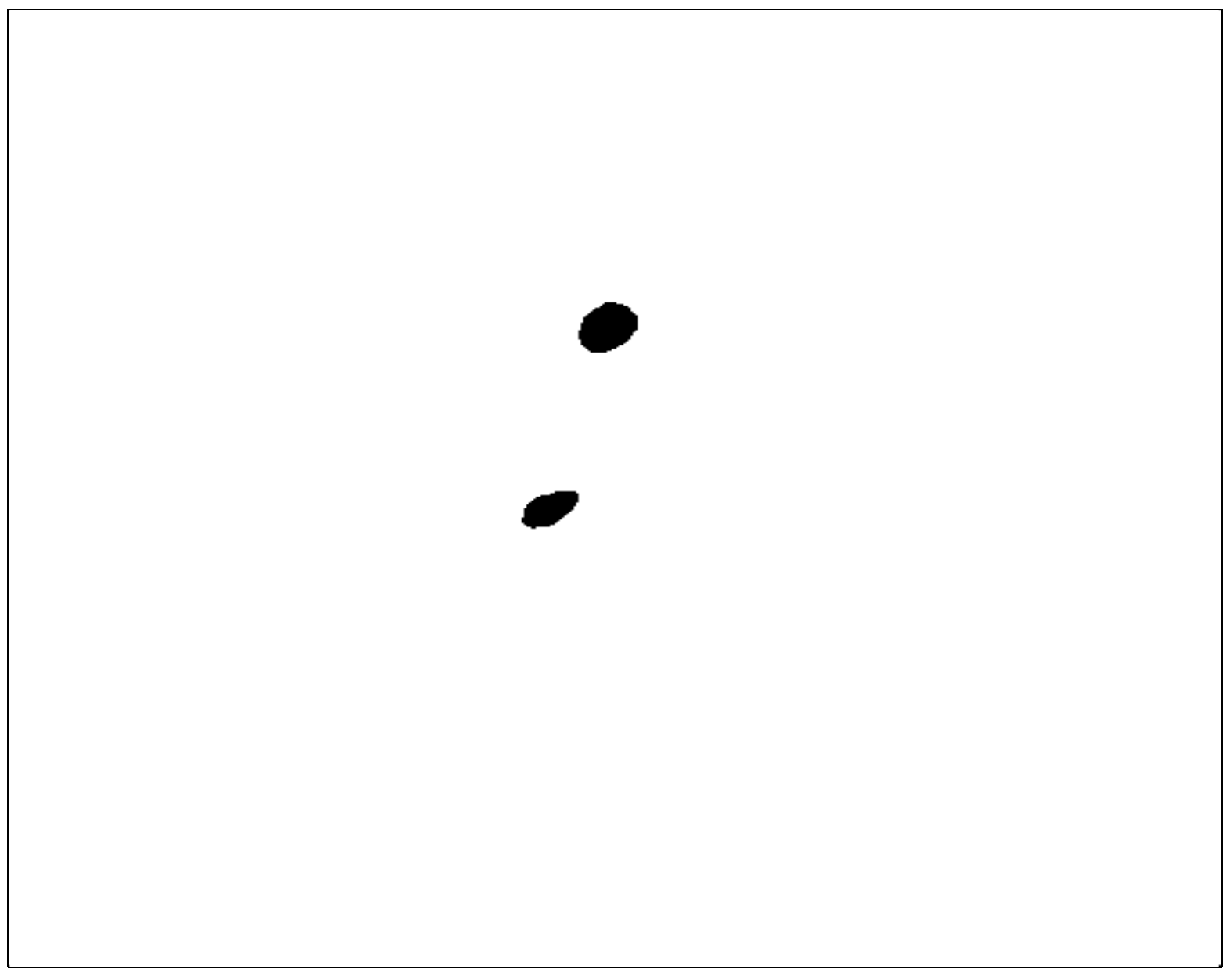}}
{\includegraphics[width=3cm,height=2.25cm]{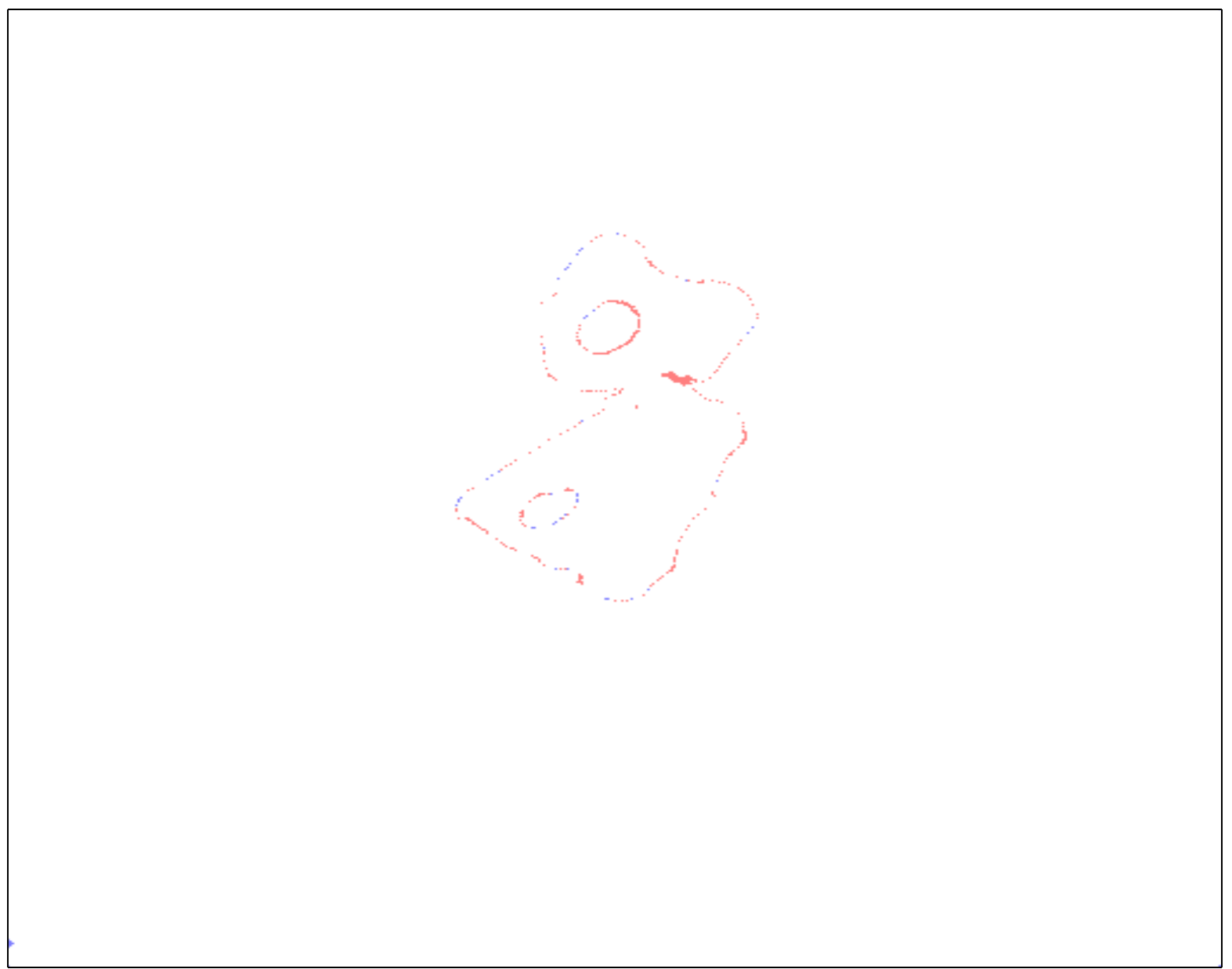}}
{\includegraphics[width=3cm,height=2.25cm]{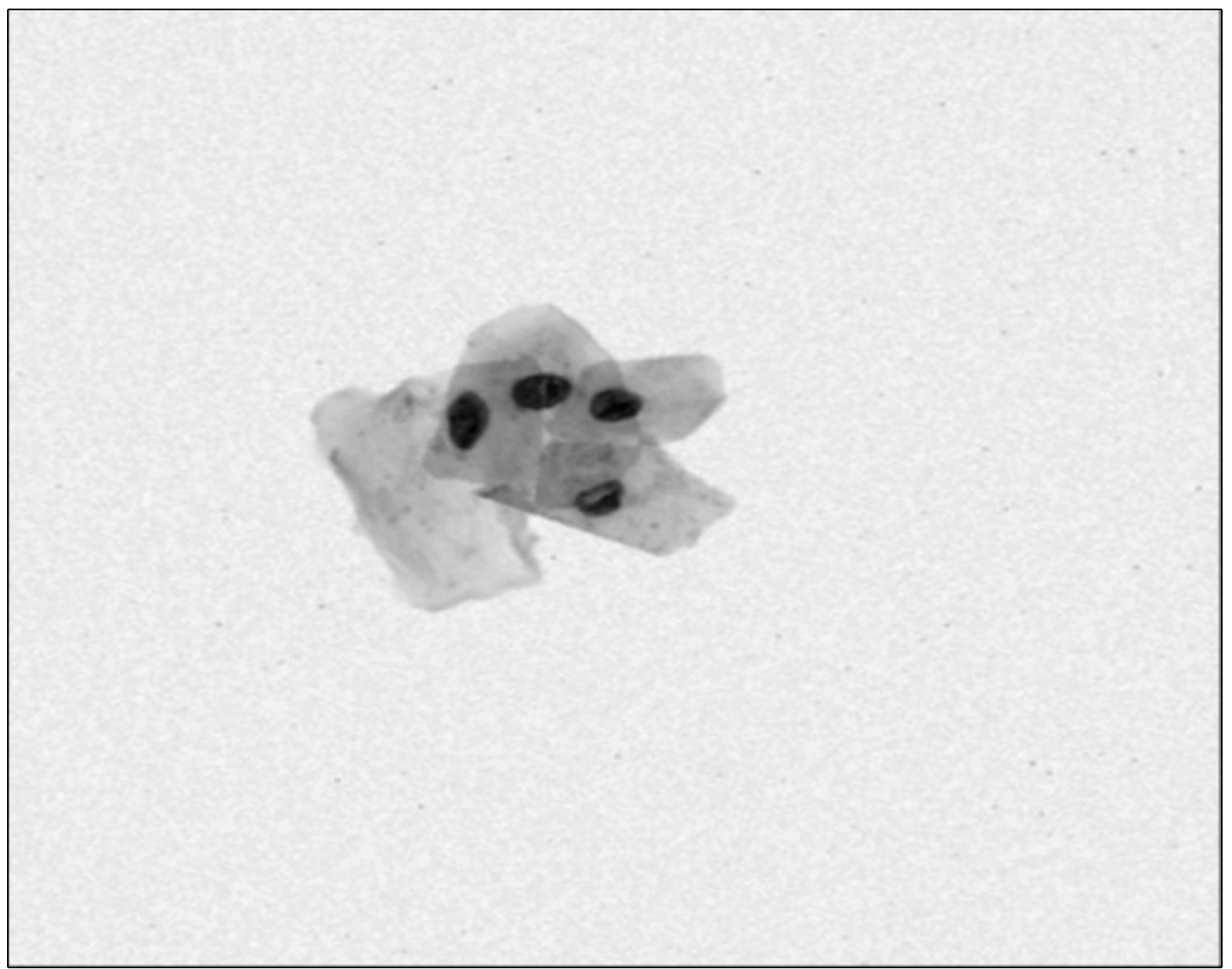}}
{\includegraphics[width=3cm,height=2.25cm]{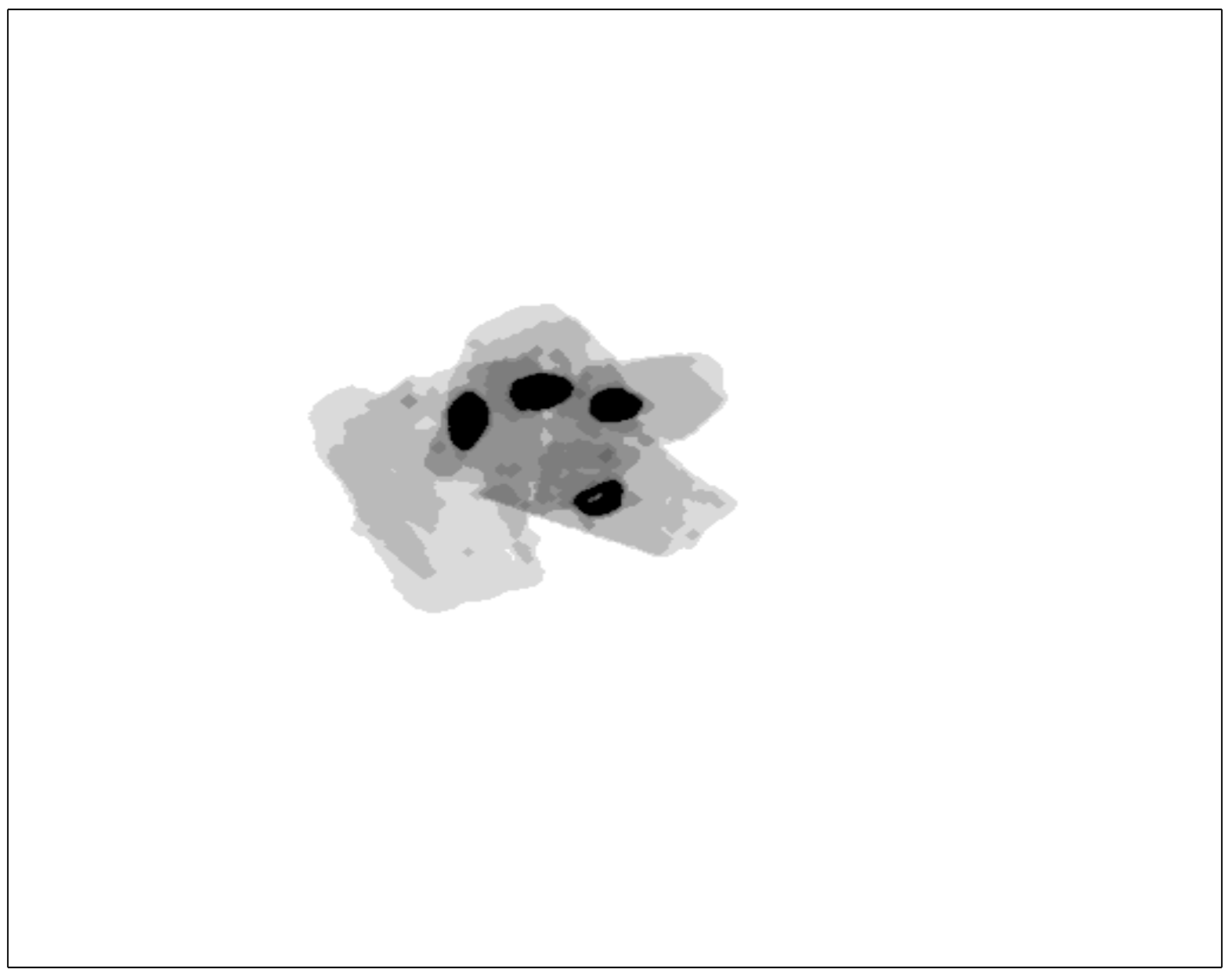}}
{\includegraphics[width=3cm,height=2.25cm]{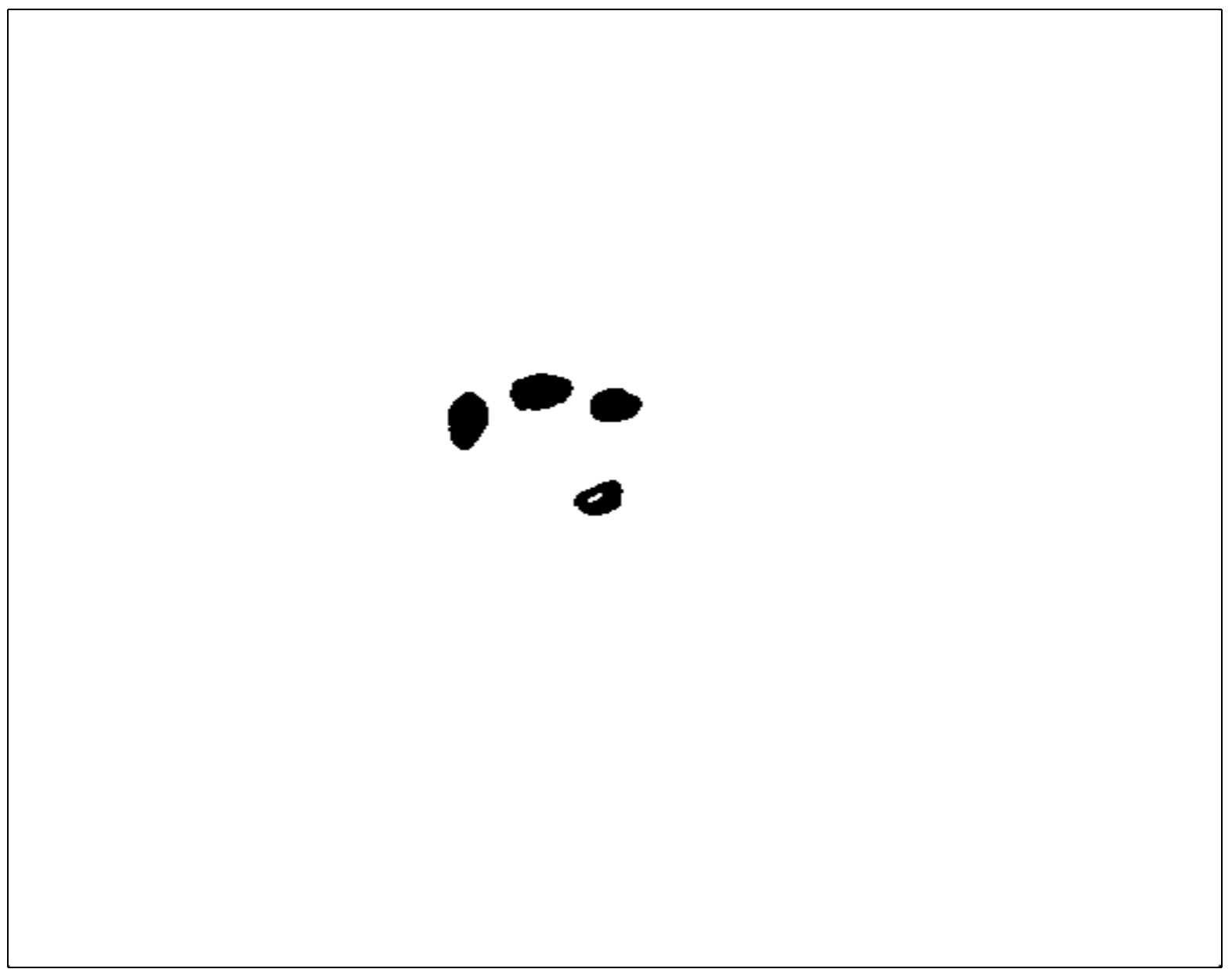}}
{\includegraphics[width=3cm,height=2.25cm]{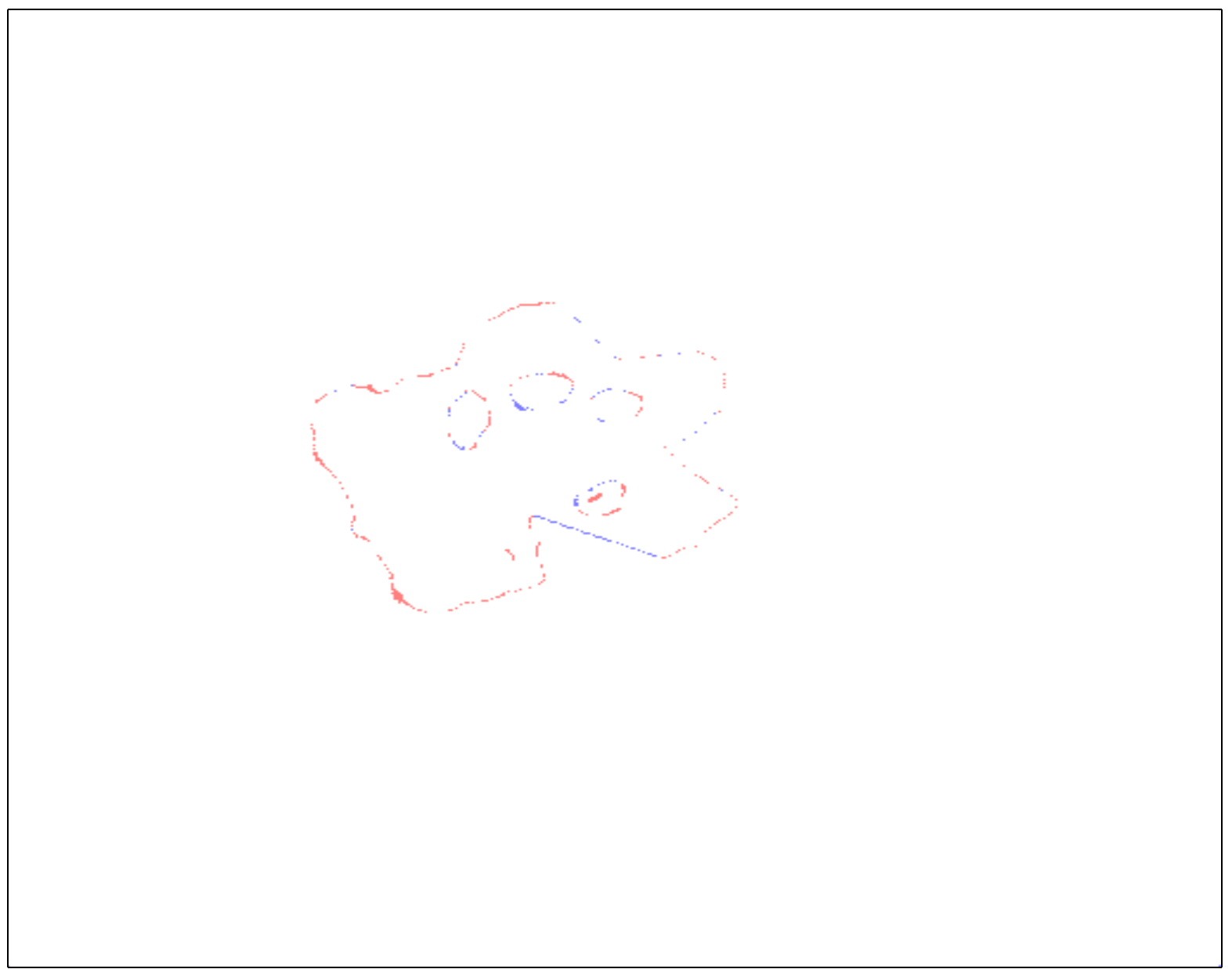}}
{\includegraphics[width=3cm,height=2.25cm]{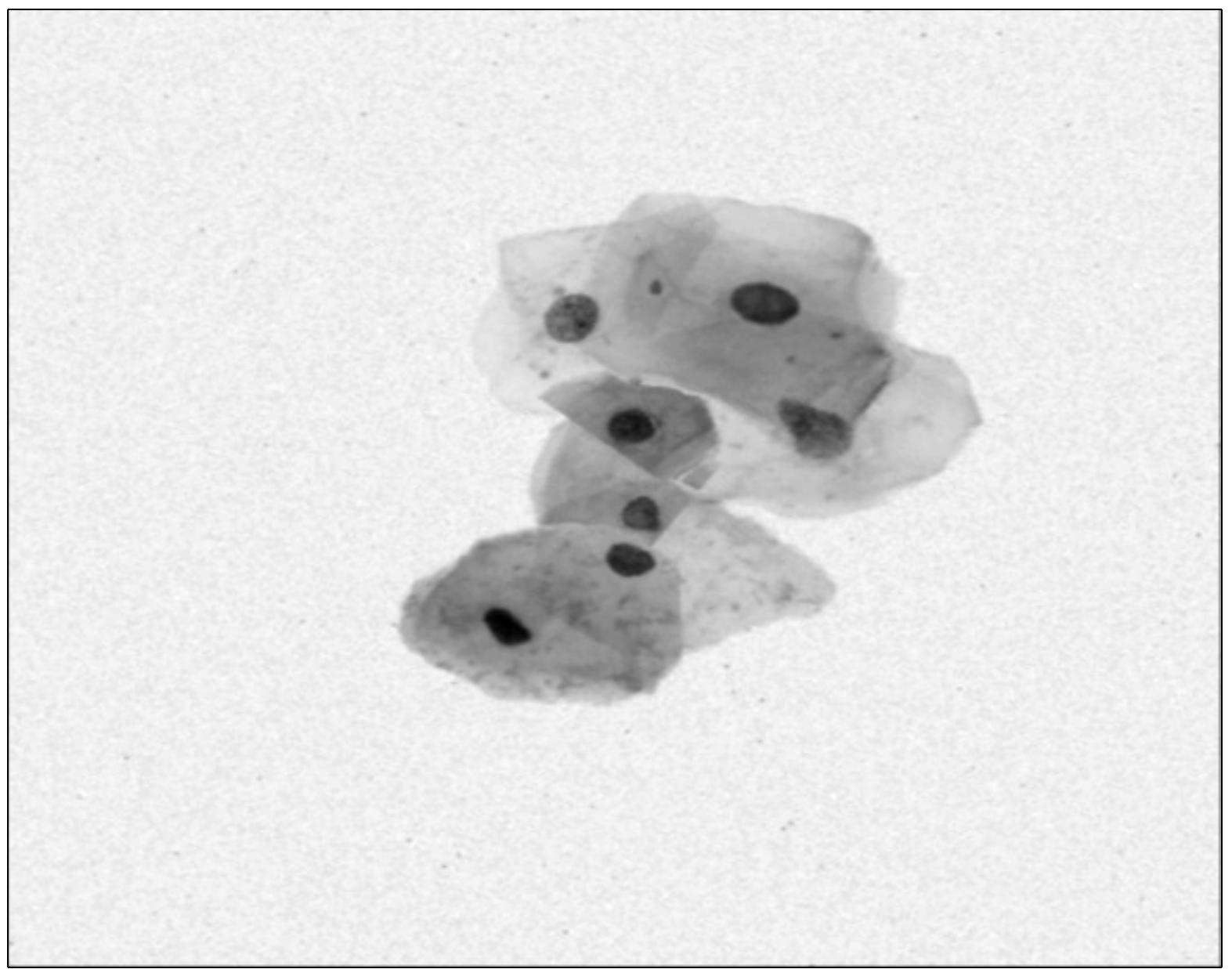}}
{\includegraphics[width=3cm,height=2.25cm]{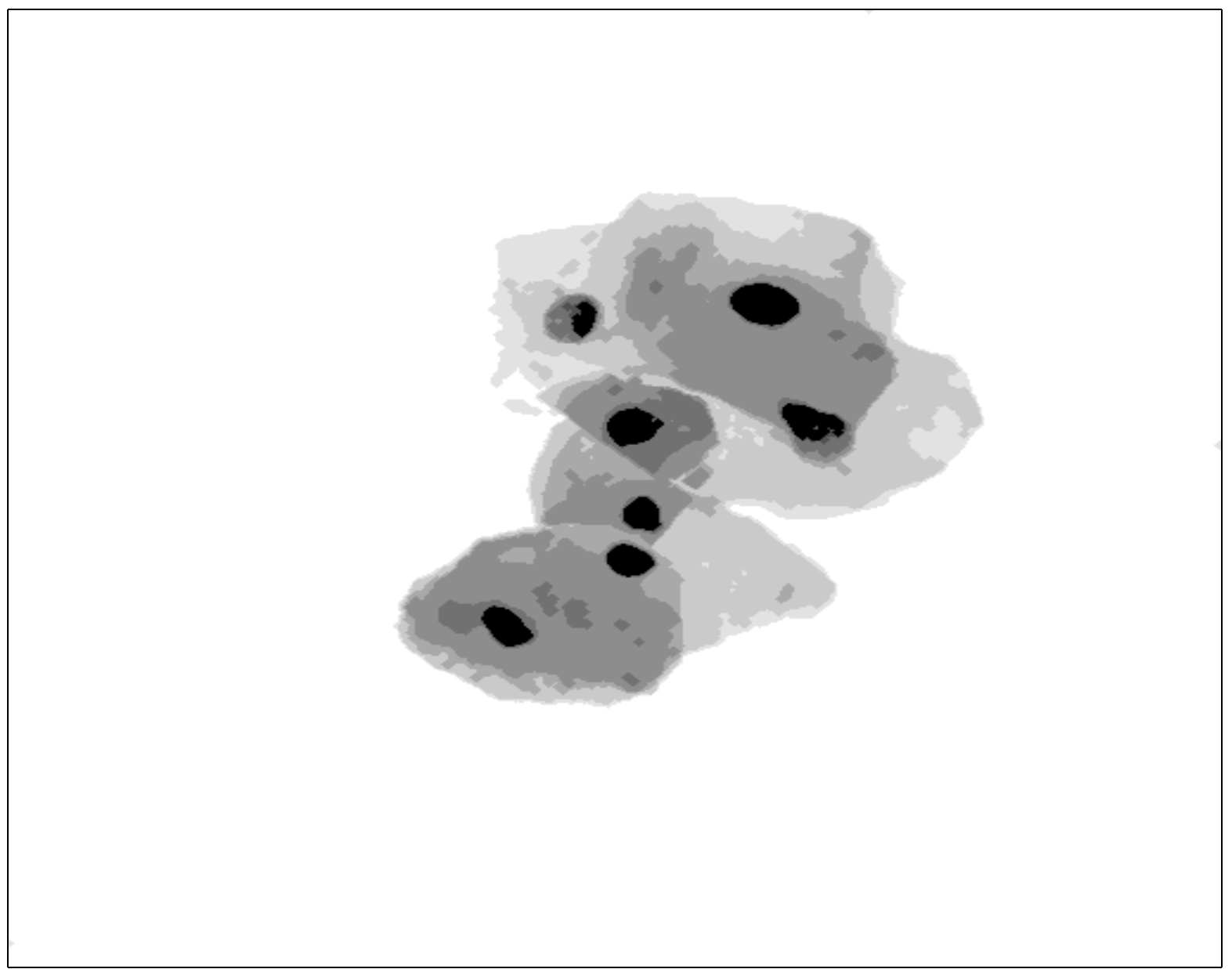}}
{\includegraphics[width=3cm,height=2.25cm]{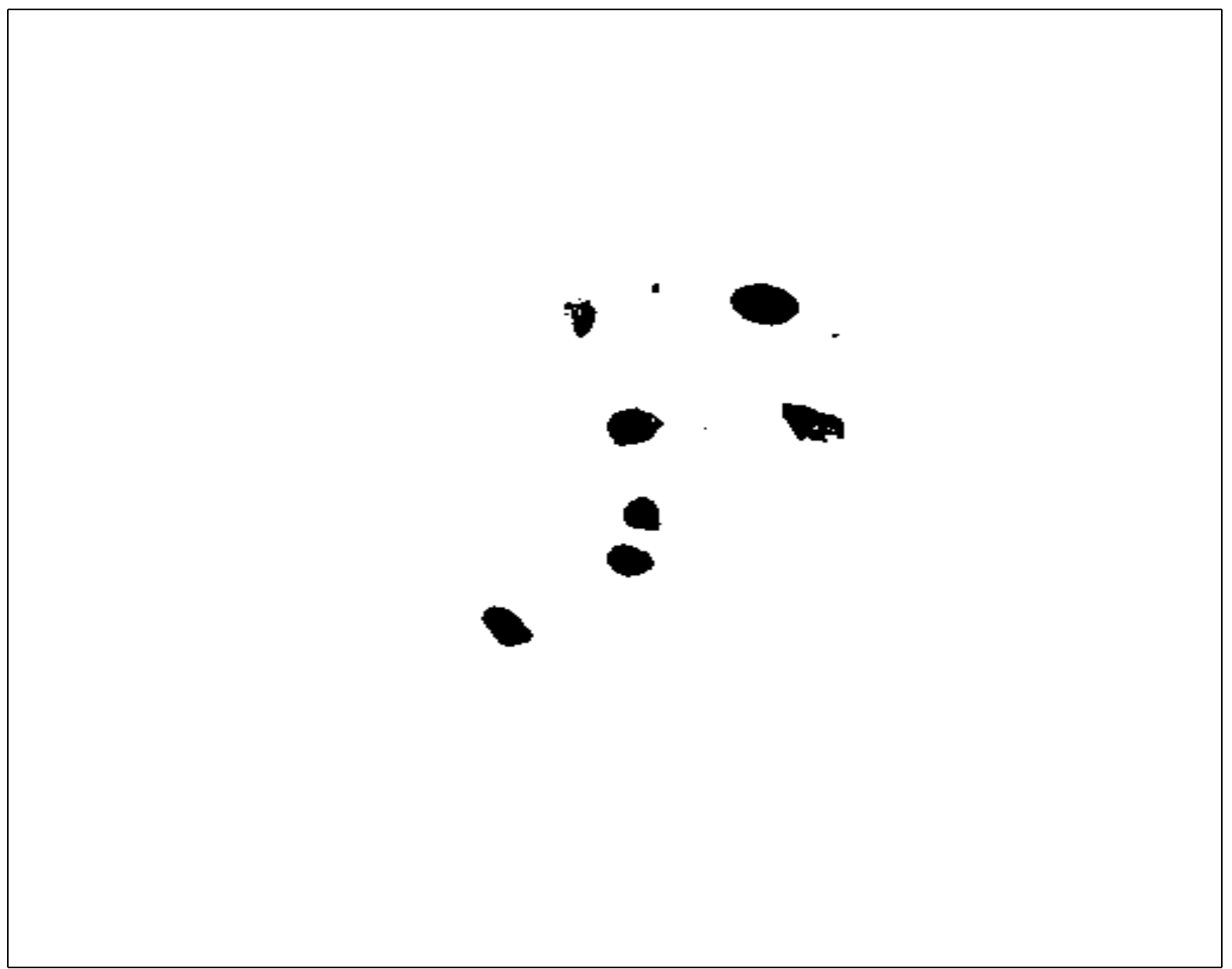}}
{\includegraphics[width=3cm,height=2.25cm]{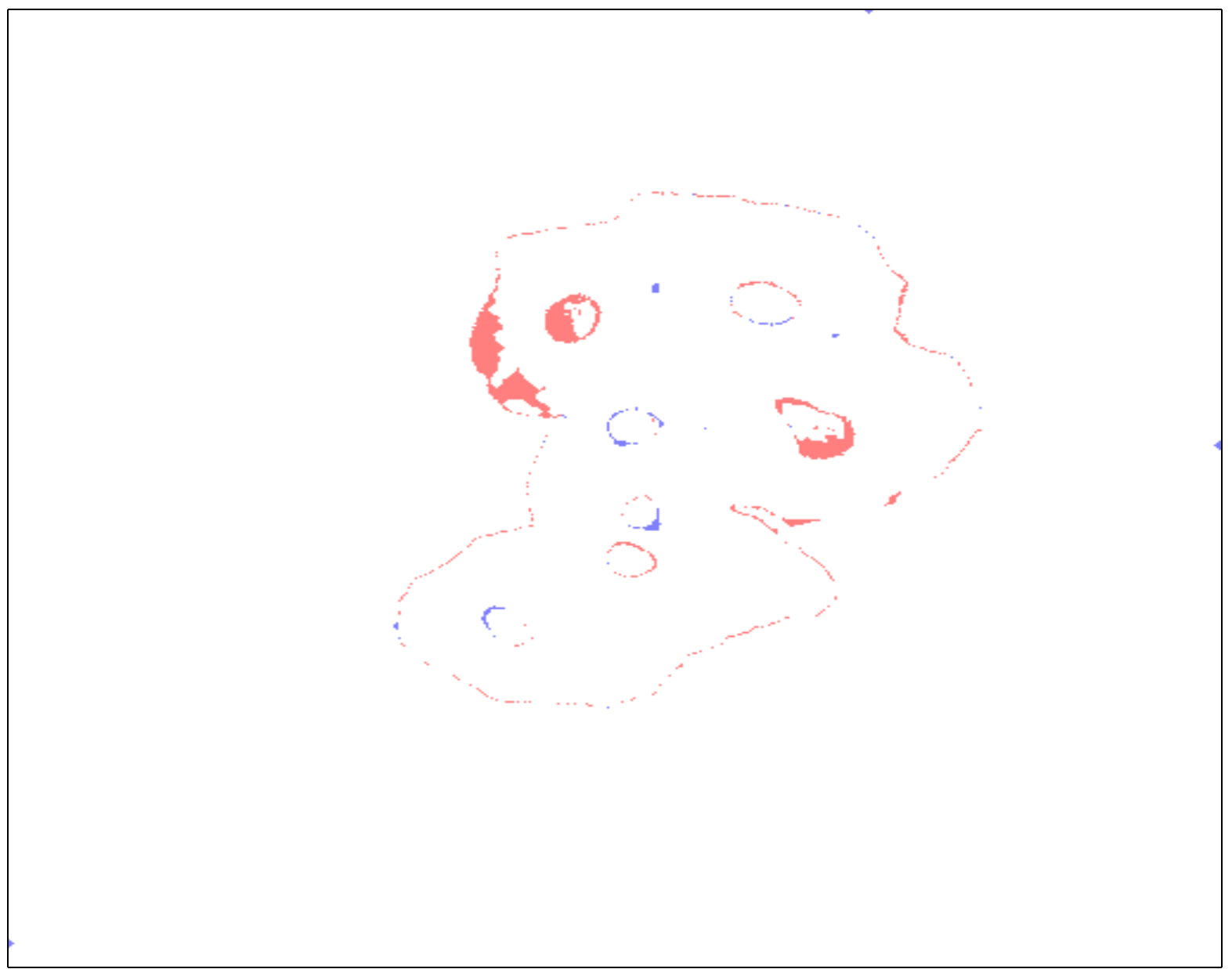}}
{\includegraphics[width=3cm,height=2.25cm]{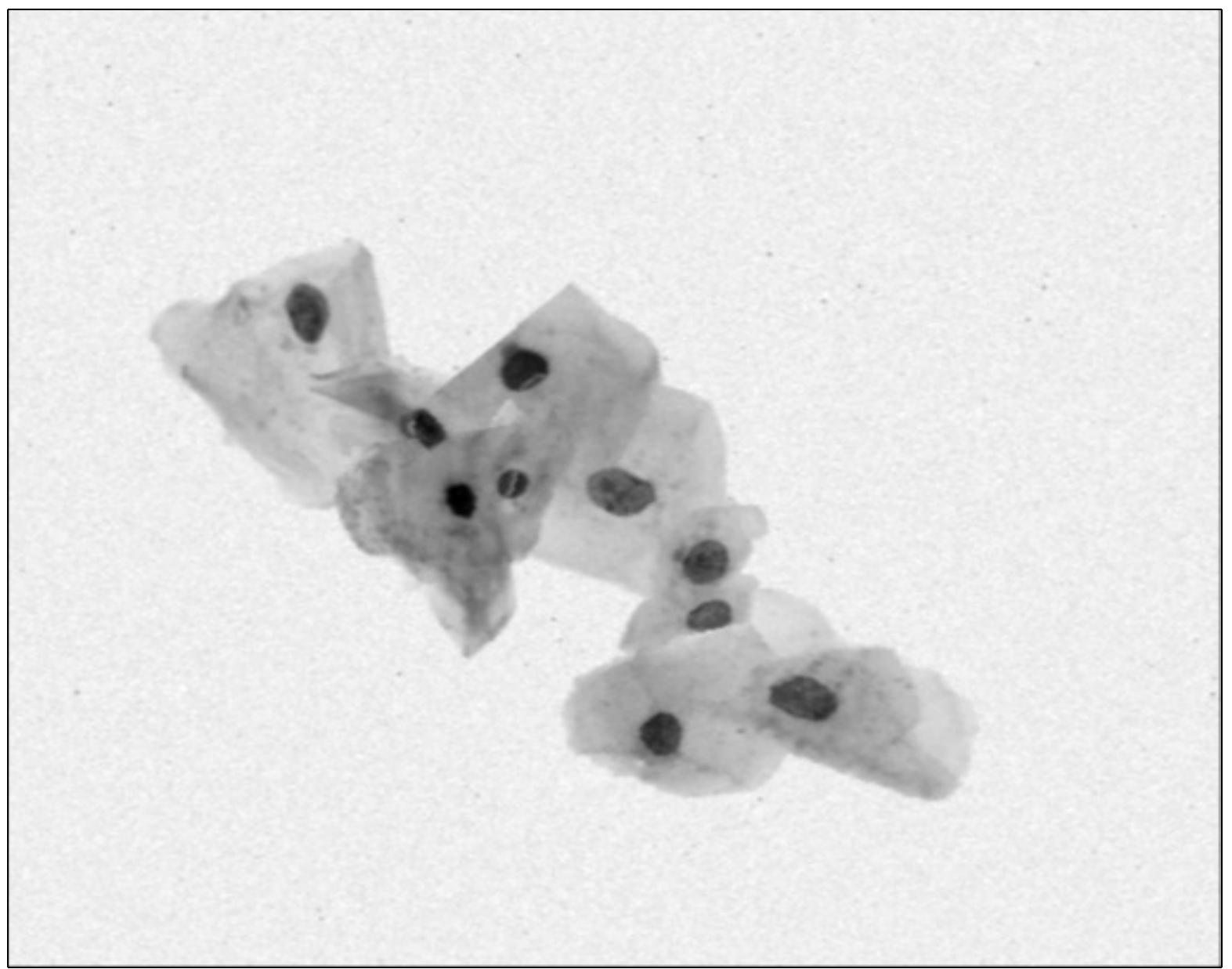}}
{\includegraphics[width=3cm,height=2.25cm]{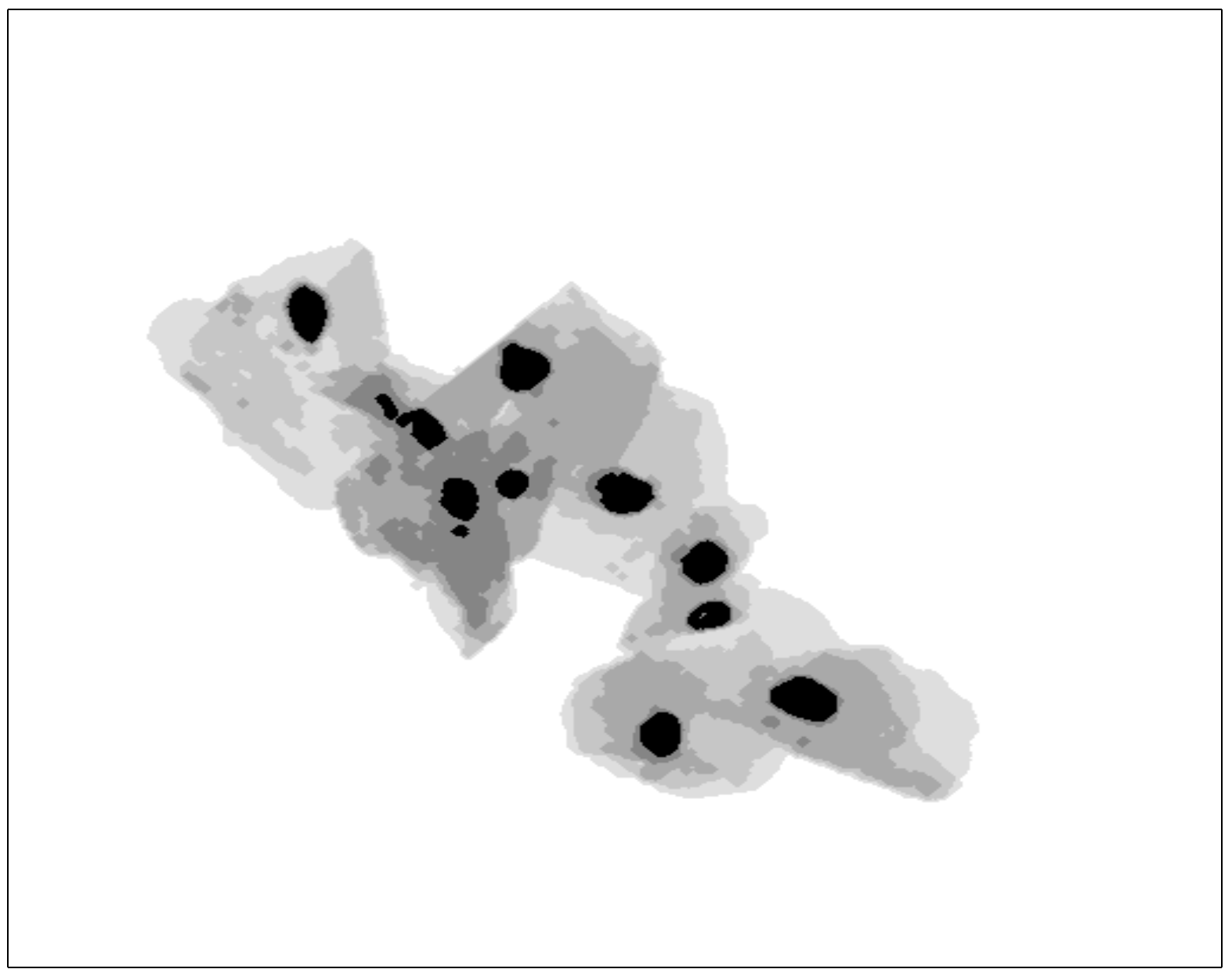}}
{\includegraphics[width=3cm,height=2.25cm]{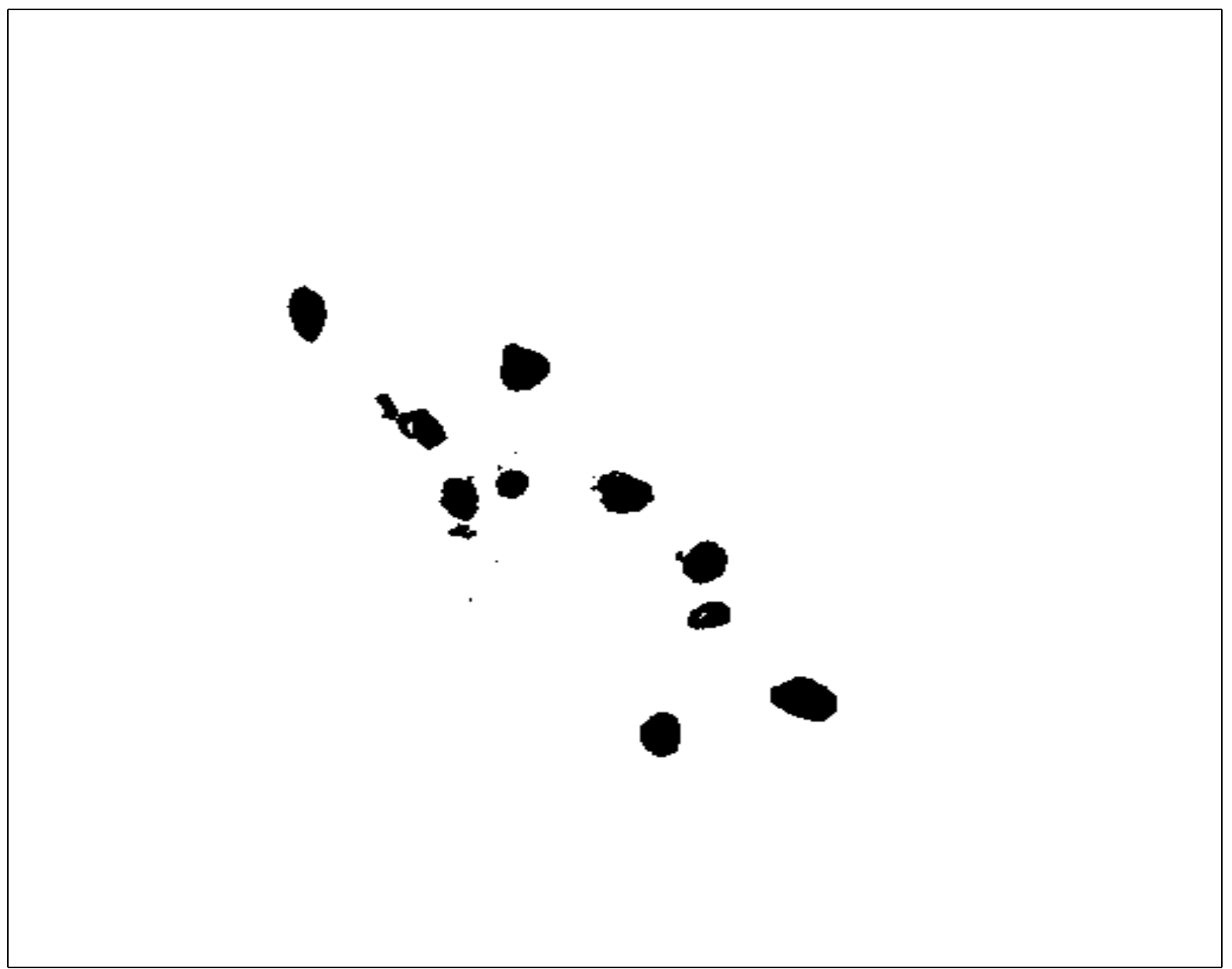}}
{\includegraphics[width=3cm,height=2.25cm]{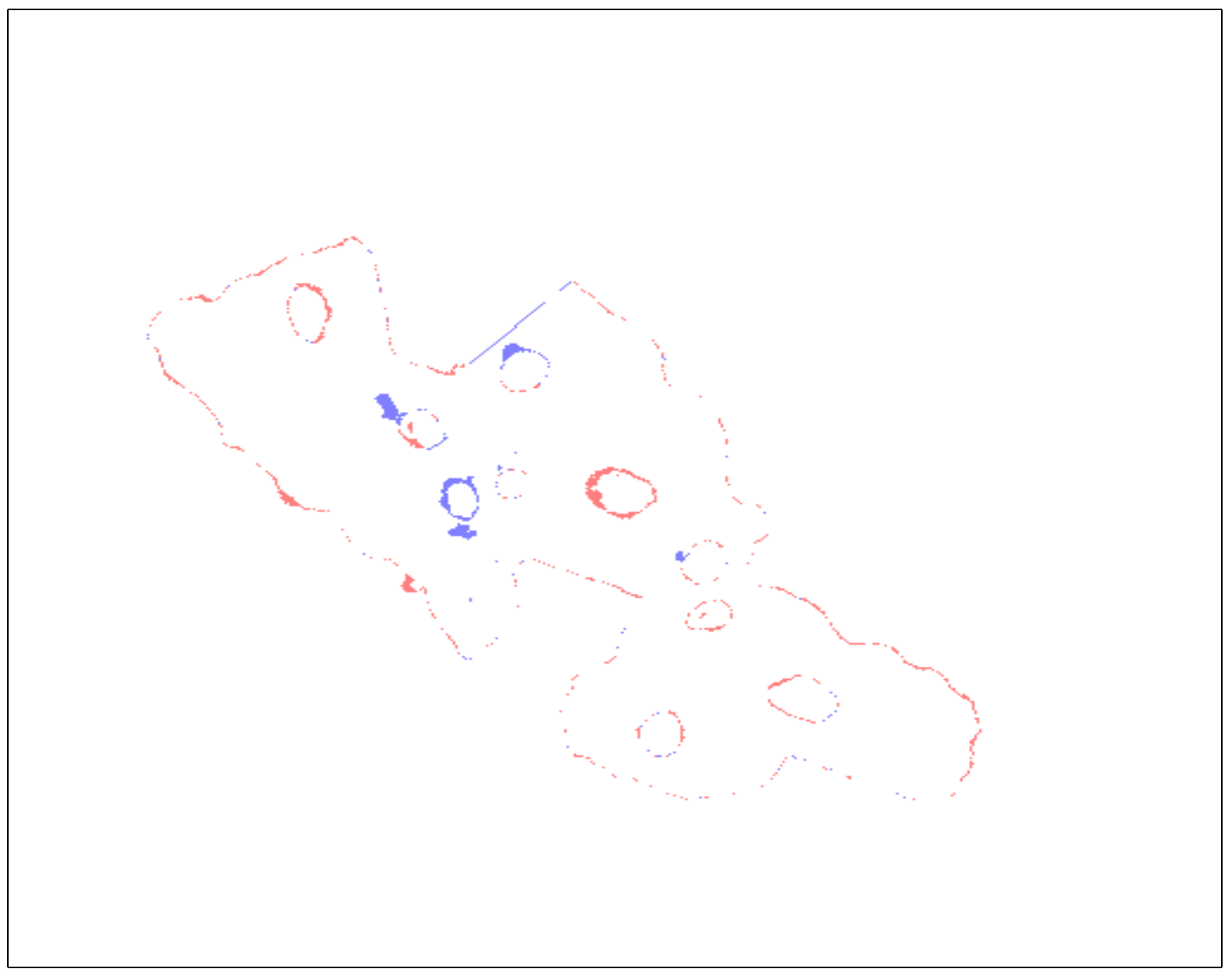}}
{\includegraphics[width=3cm,height=2.25cm]{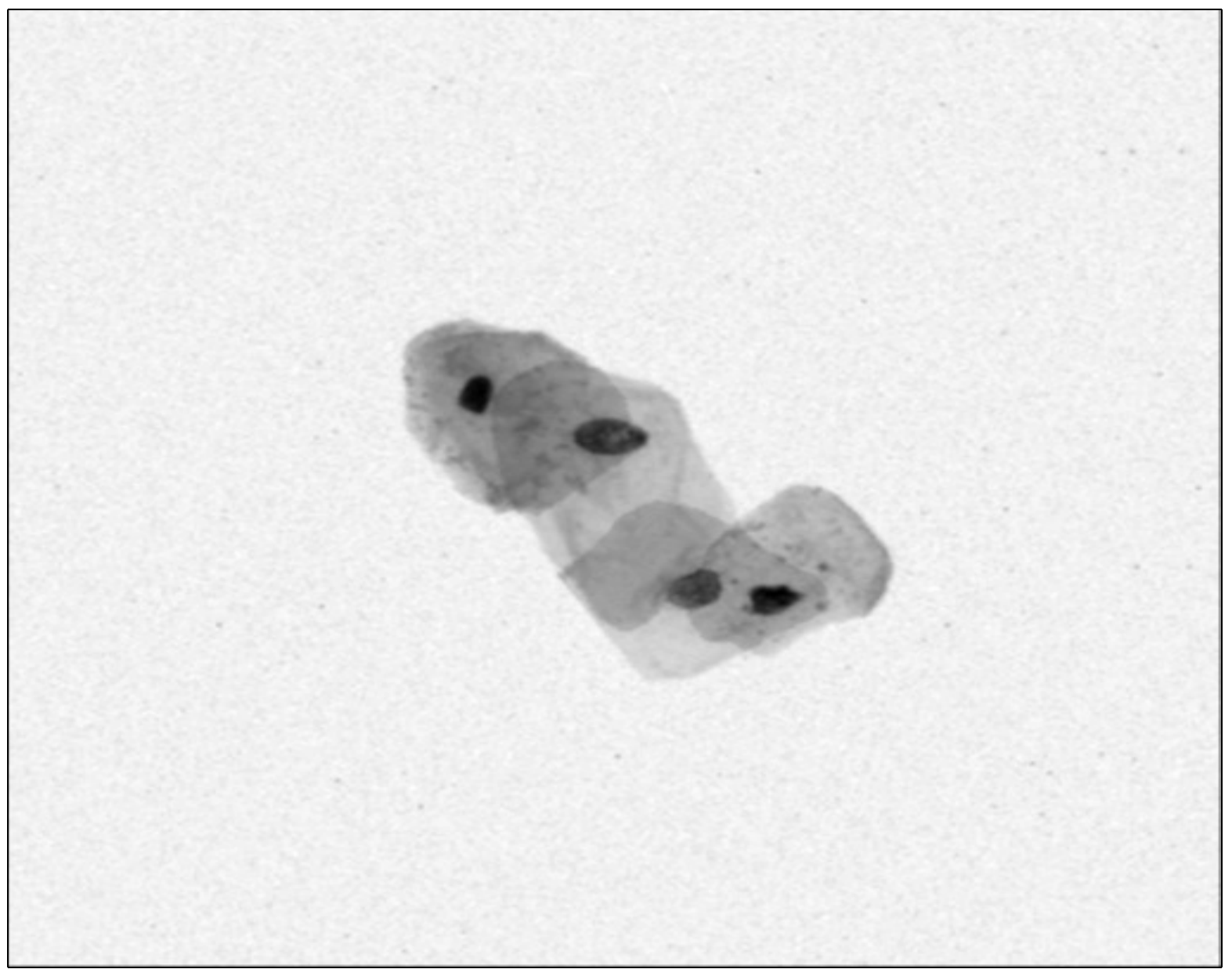}}
{\includegraphics[width=3cm,height=2.25cm]{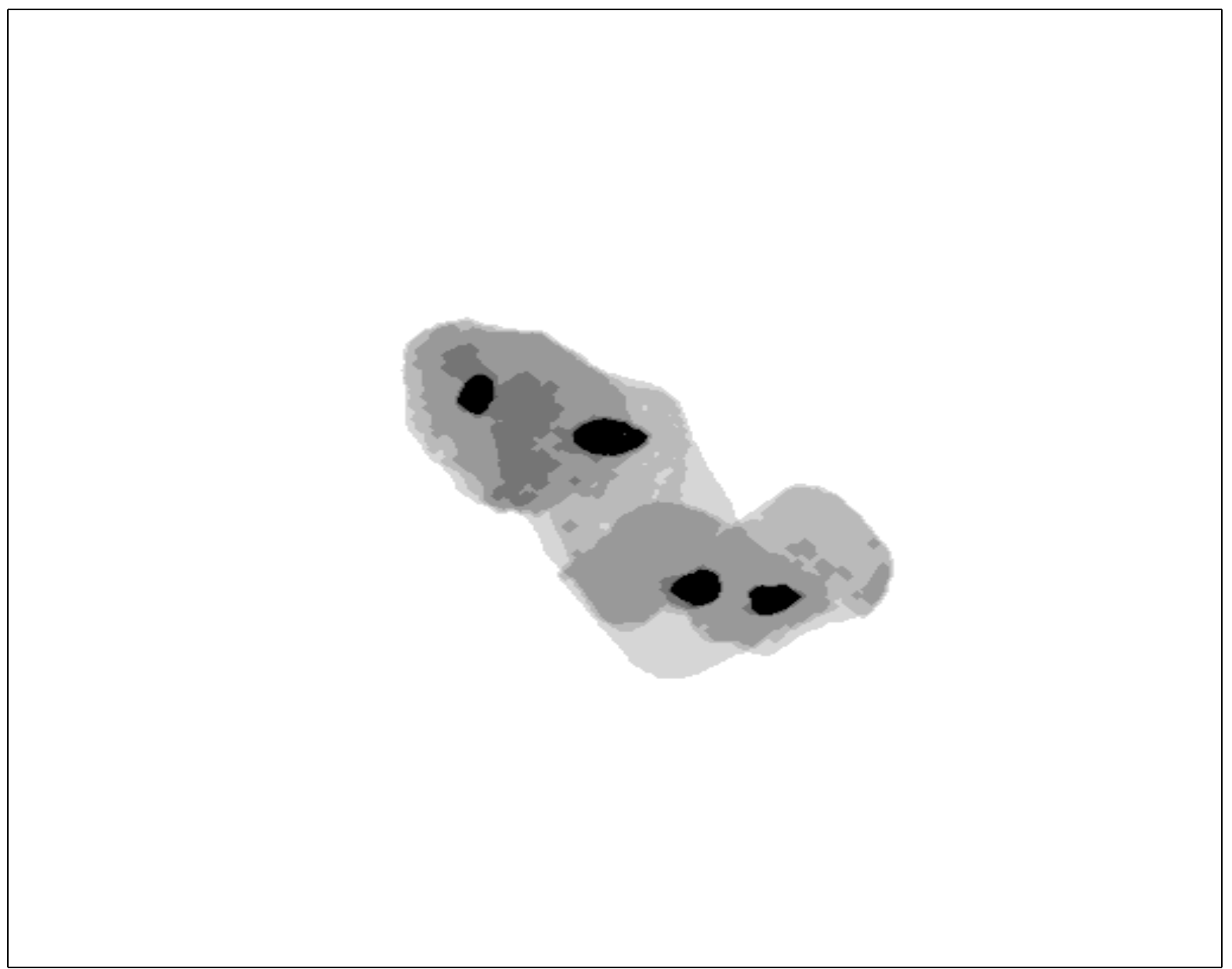}}
{\includegraphics[width=3cm,height=2.25cm]{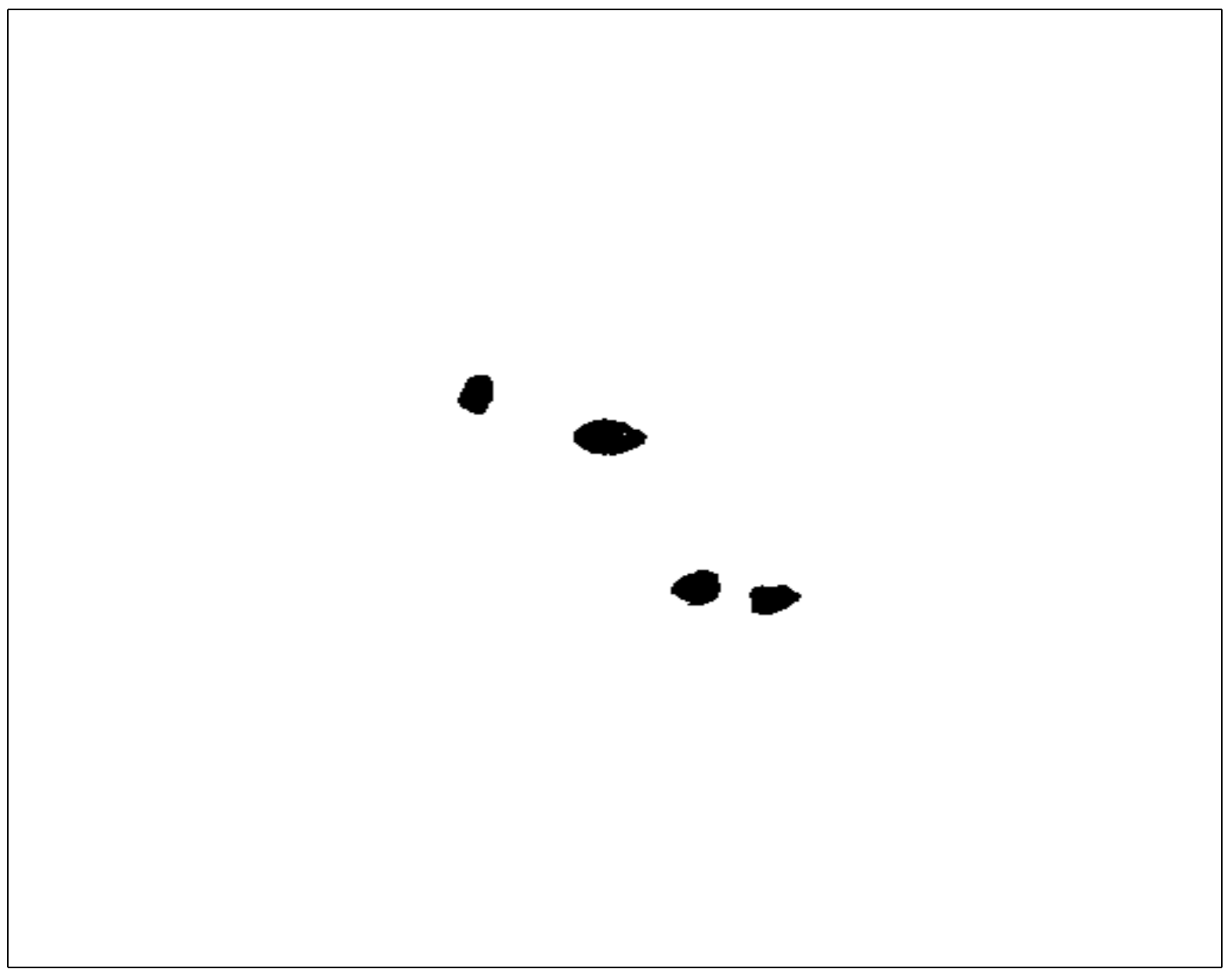}}
{\includegraphics[width=3cm,height=2.25cm]{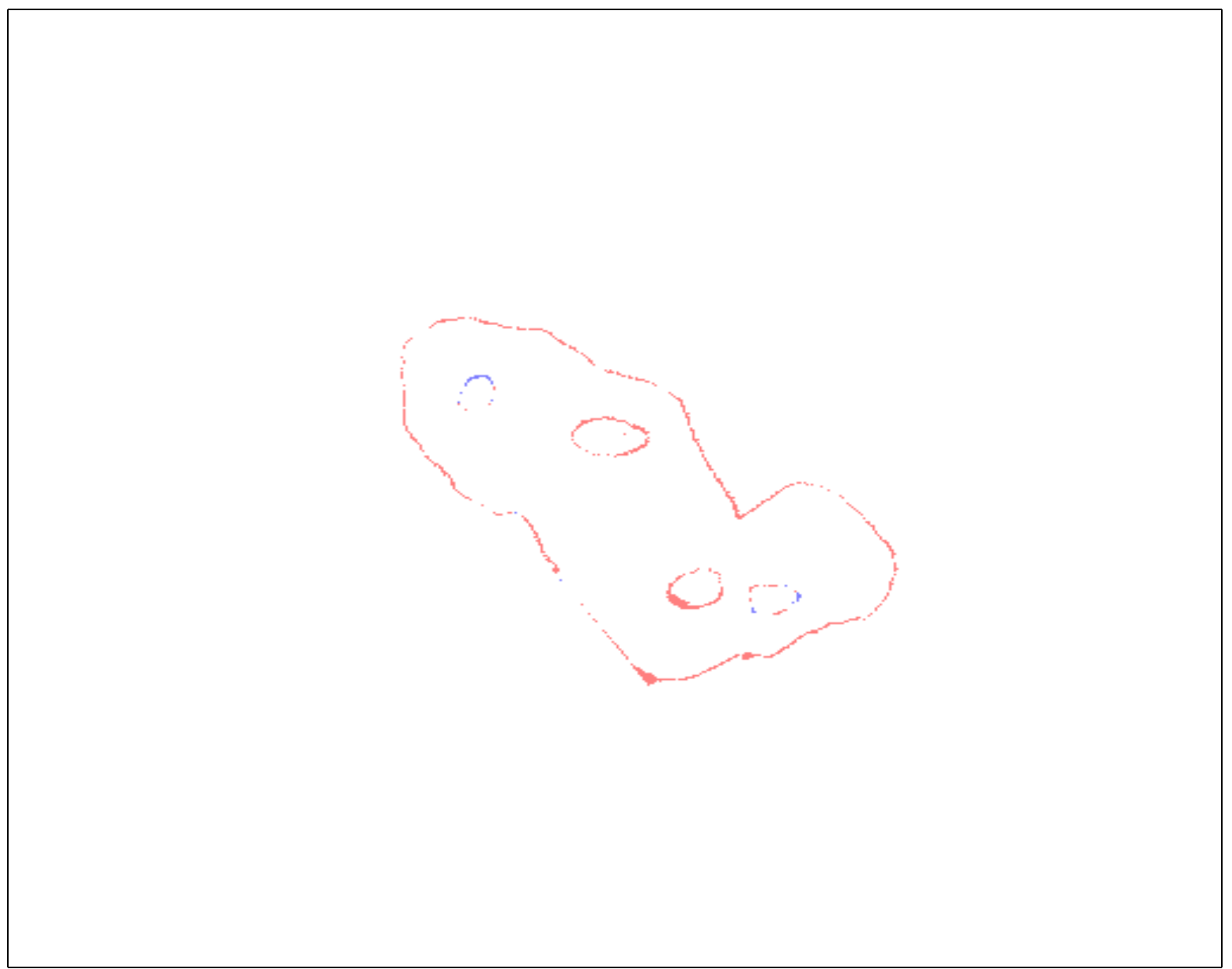}}
{\includegraphics[width=3cm,height=2.25cm]{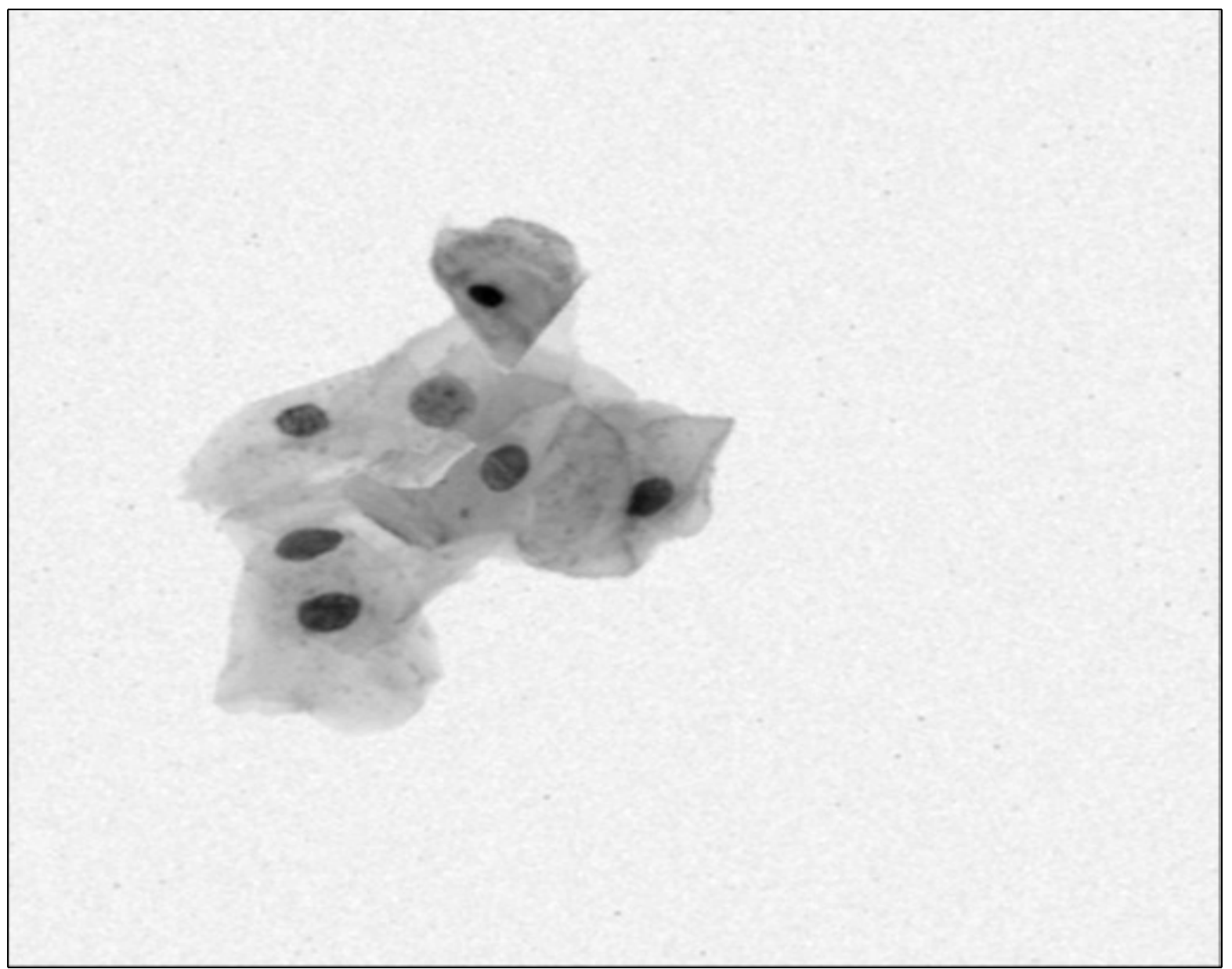}}
{\includegraphics[width=3cm,height=2.25cm]{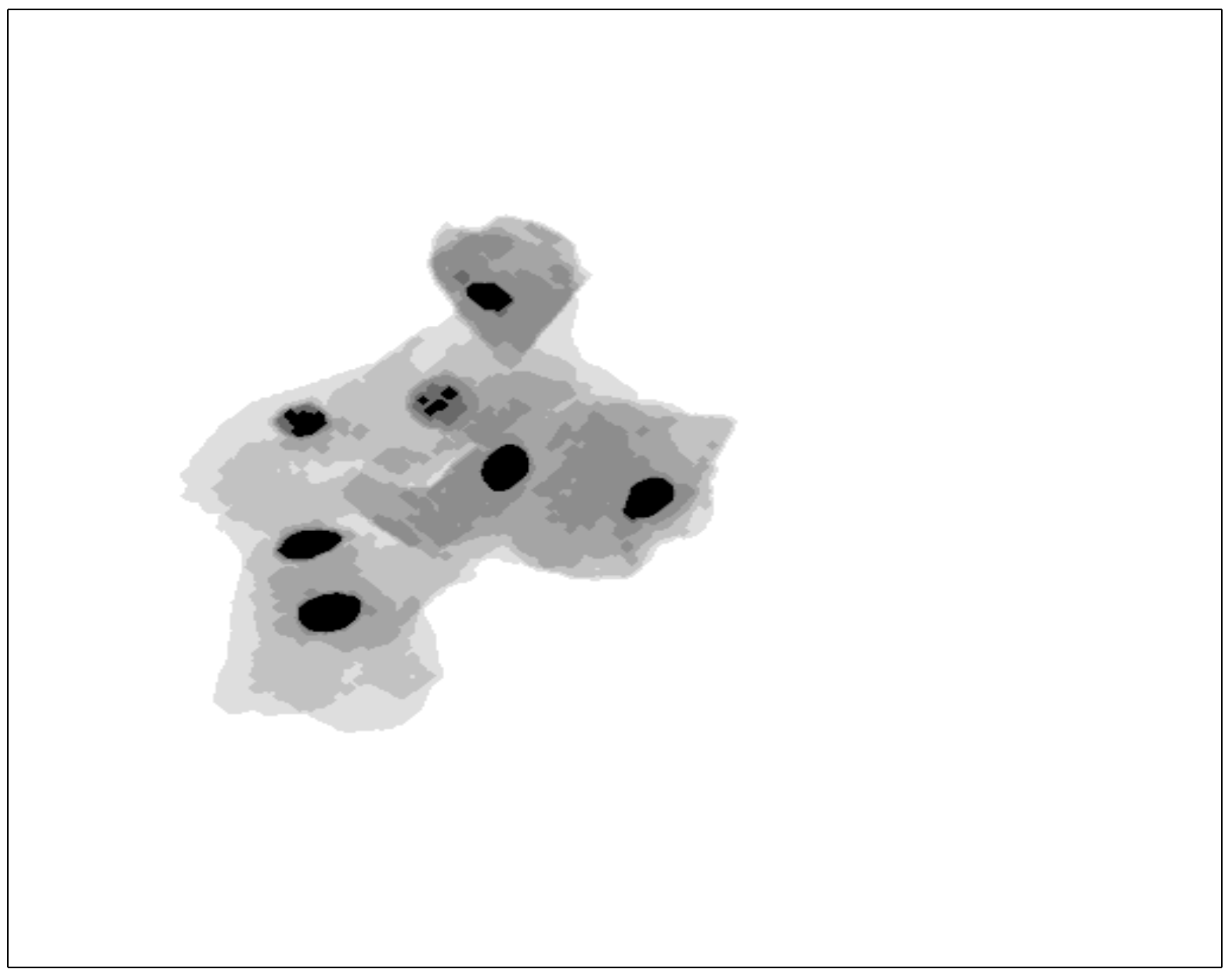}}
{\includegraphics[width=3cm,height=2.25cm]{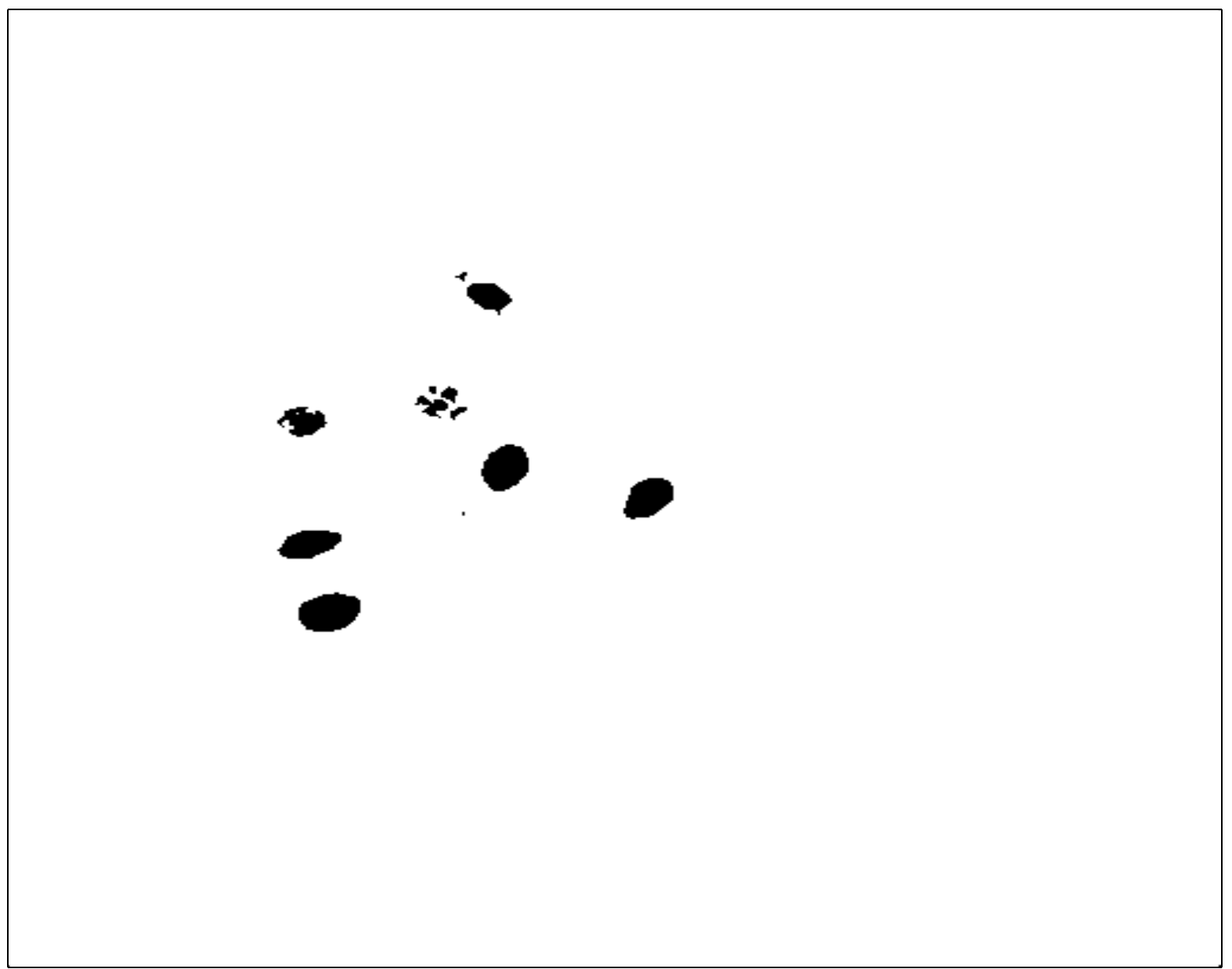}}
{\includegraphics[width=3cm,height=2.25cm]{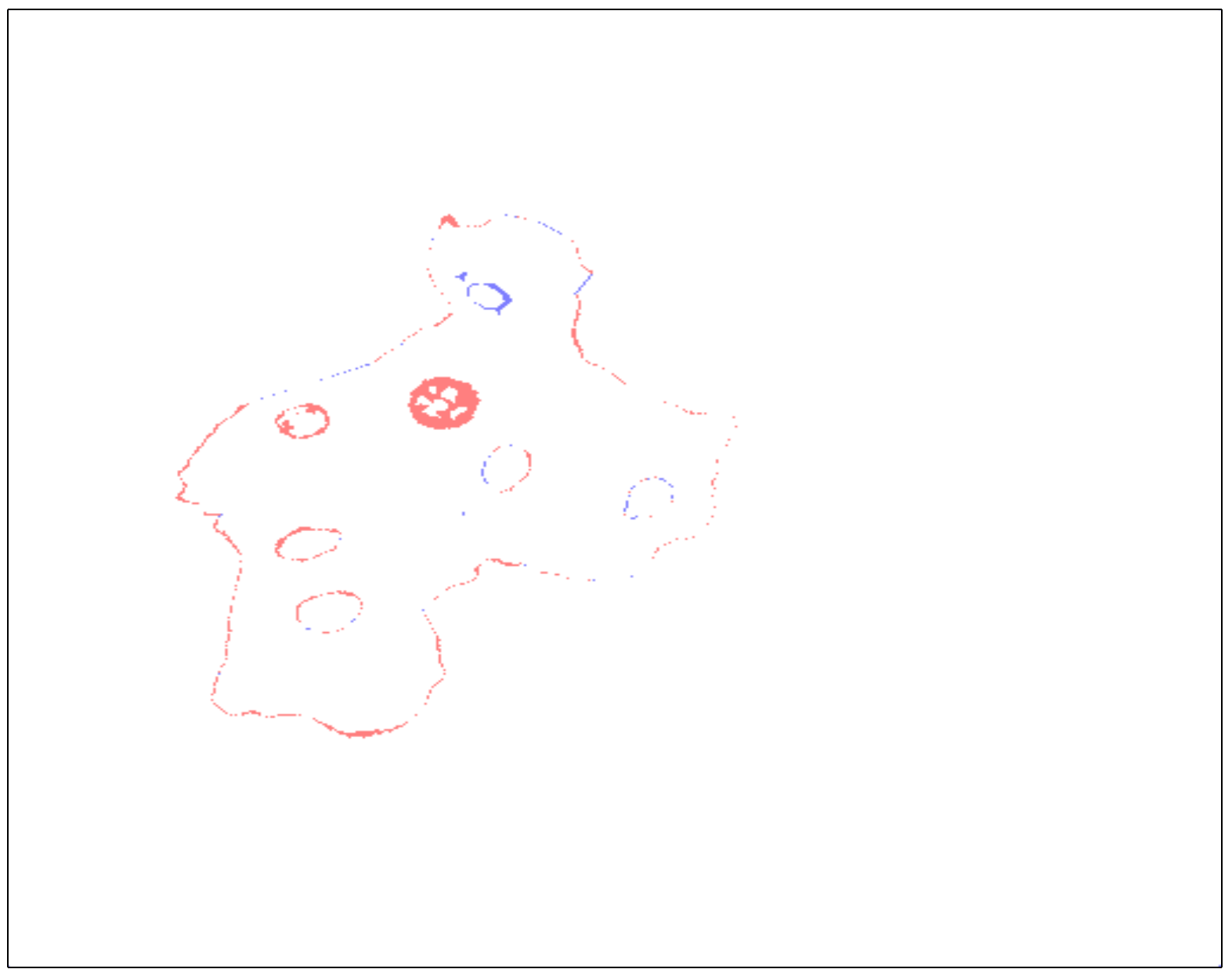}}
{\includegraphics[width=3cm,height=2.25cm]{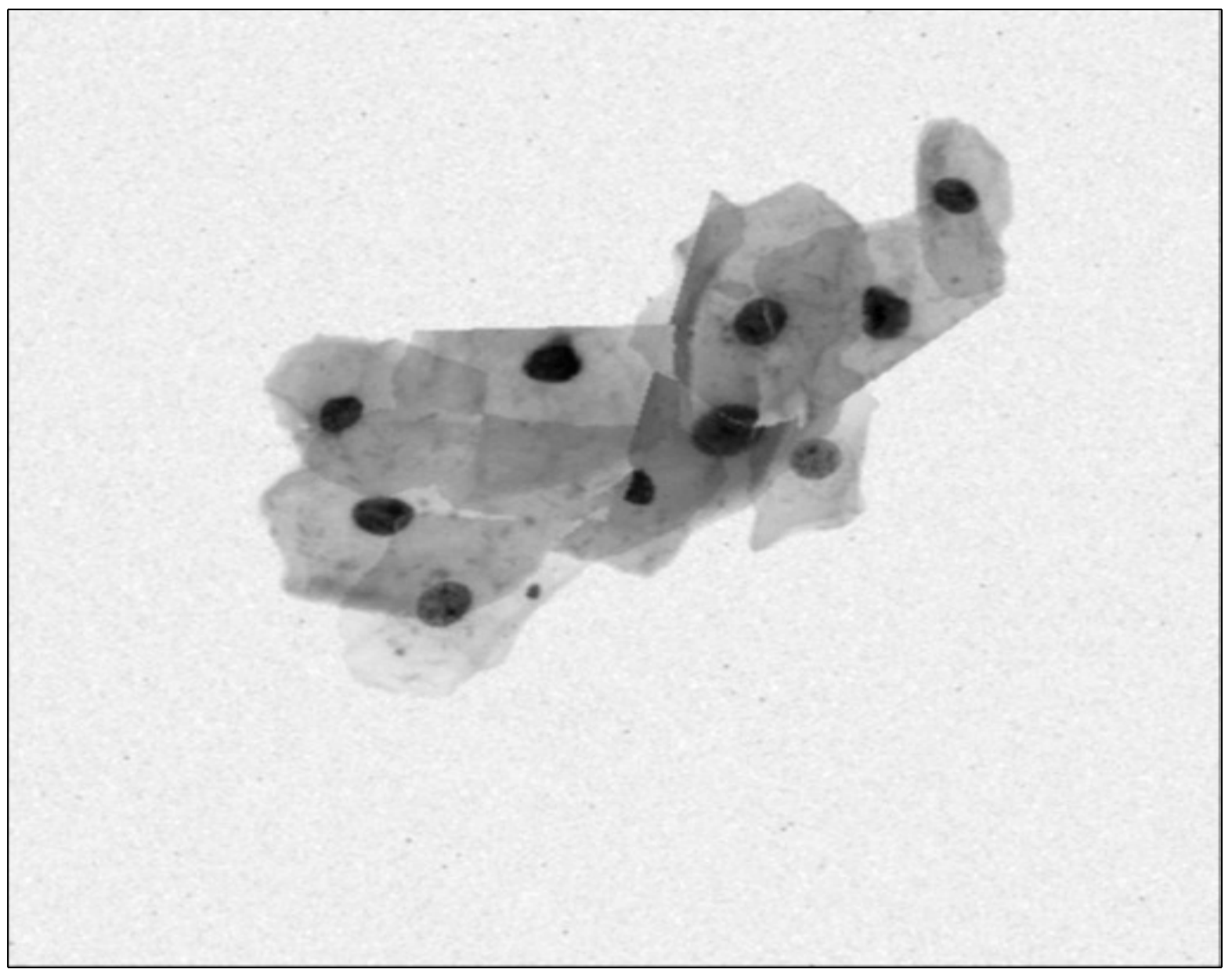}}
{\includegraphics[width=3cm,height=2.25cm]{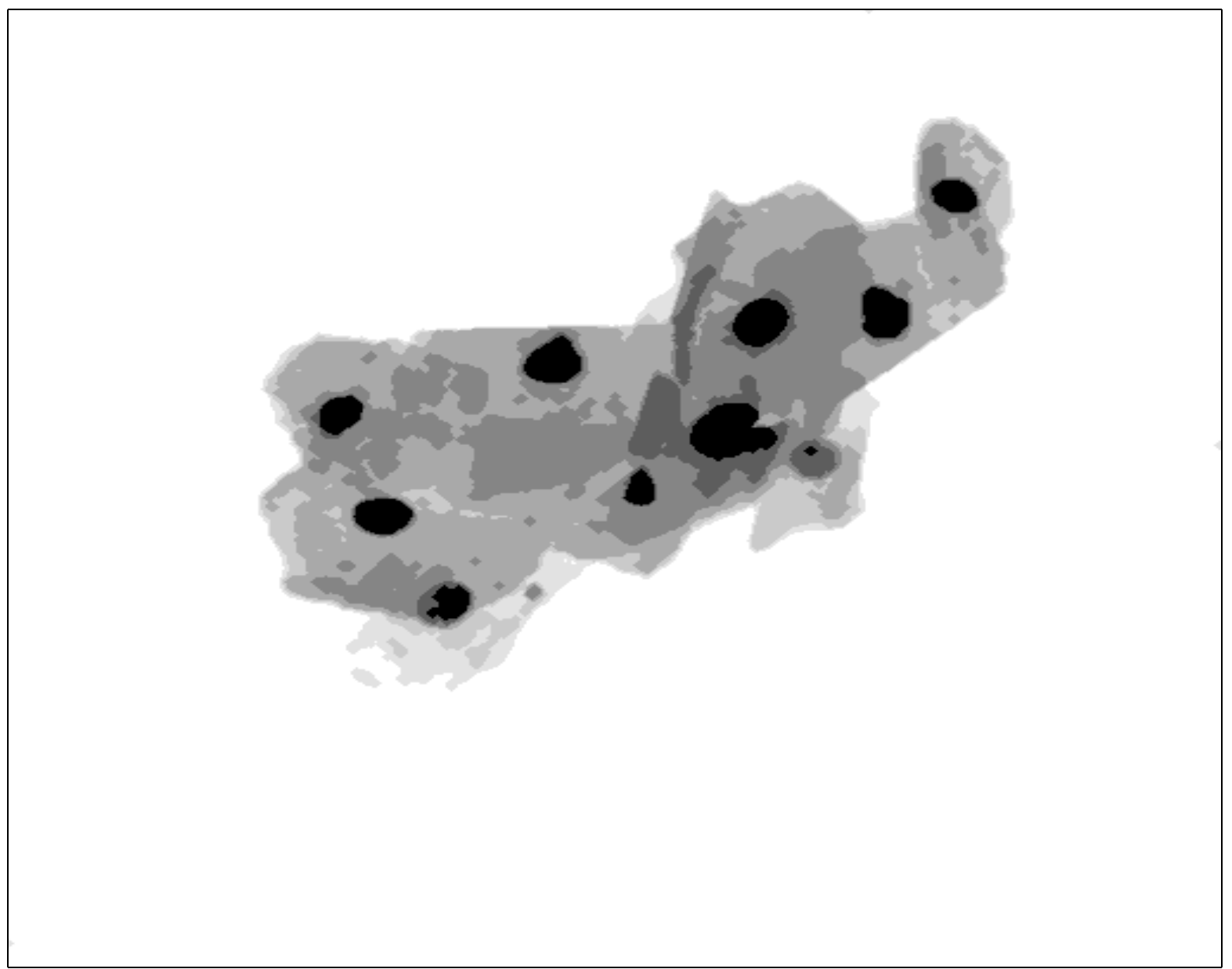}}
{\includegraphics[width=3cm,height=2.25cm]{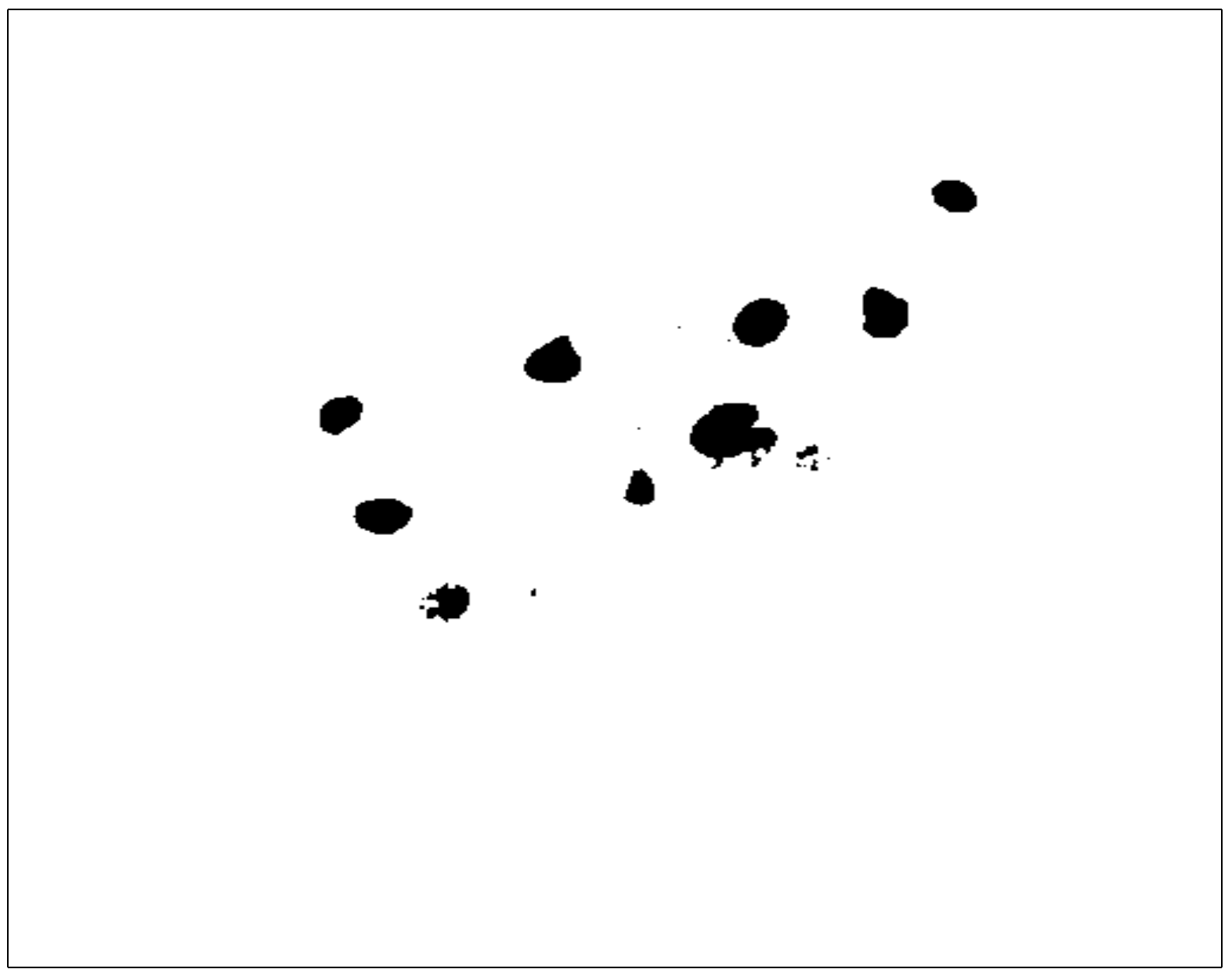}}
{\includegraphics[width=3cm,height=2.25cm]{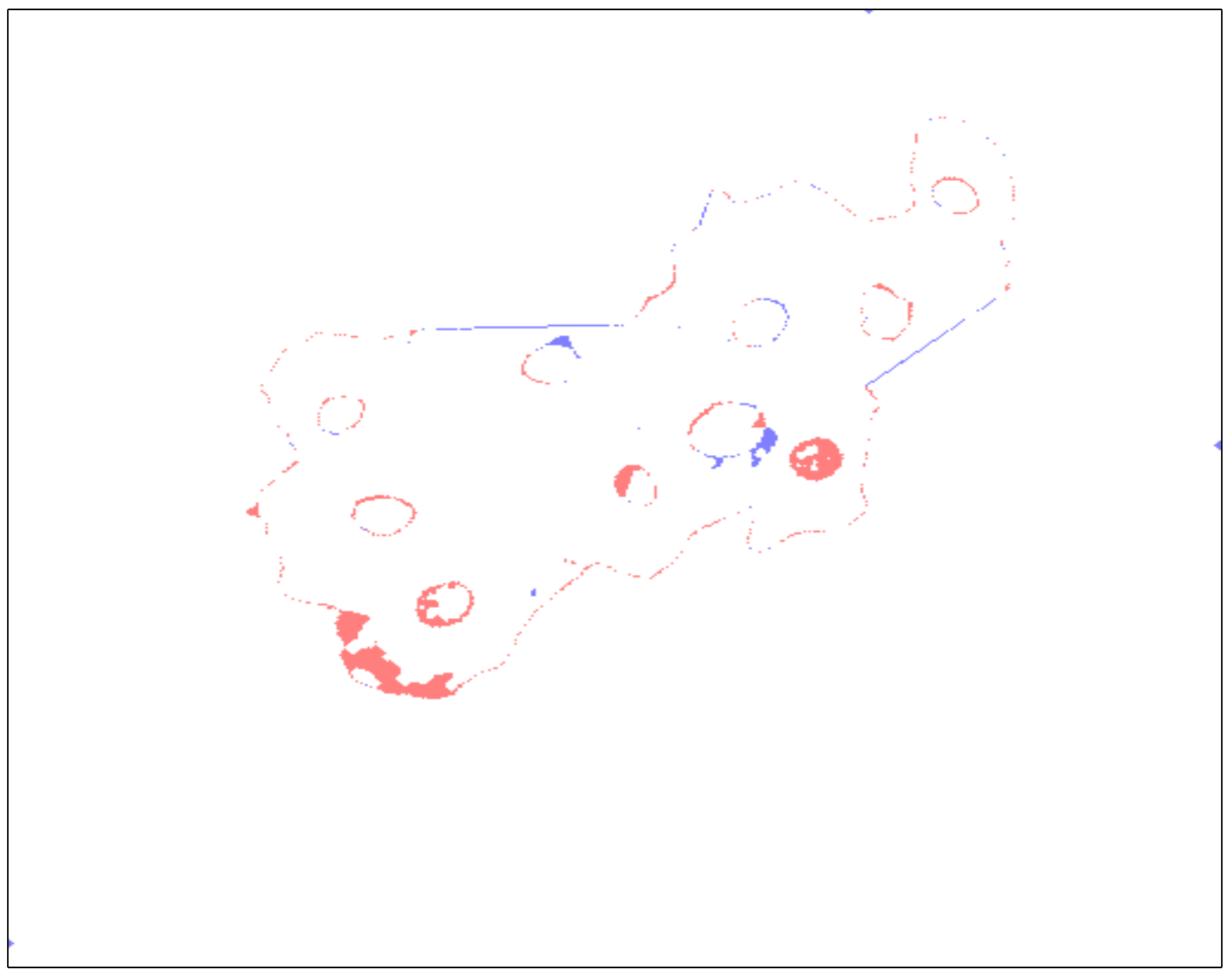}}
\caption{ Example. Nucleus and citoplasm segmentation process. First column corresponds to the initial image. Second column, to the background extraction, and third column to the nucleus segmentation. The citoplasm is the difference between the images shown 
in the third and second columns. Finally, fourth column shows the difference between the ground-truth nucleus segmentation and the obtained with our method. }
\label{fig1} 
\end{figure}

In Figure~\ref{fig1}, we show the segmentation process for the two regions of interest. The first column corresponds to the initial image. The second column, to the background extraction, and the third column to the nucleus segmentation. Thus, the cytoplasm is the difference between the images shown in the third and second column. 
Finally, the fourth column shows the difference between the ground-truth nucleus segmentation and the obtained with our method.

\begin{table}[t] 
\caption{Example: Segmentation results for some samples of test90 dataset.  } 
\begin{tabular}{|l||c|c|c|c||c|c|c|c|} 
\hline\hline 
Sample &    1 &    15 &    30 &   45 &    60 &   75 &    90  & All (mean)\tabularnewline \hline 
Cytoplasm DC & 0.99 &  0.99 &  0.99 & 0.99 &  0.99 & 0.99 &  0.99  & 0.98 \tabularnewline \hline 
Nucleus DC & 0.93 &  0.94 &  0.82 & 0.88 &  0.90 & 0.81 &  0.85  & 0.87\tabularnewline \hline 
Execution time & 2.44 &  4.33 &  5.11 & 5.12 &  5.57 & 5.94 &  4.18  & 5.33\tabularnewline \hline 
\hline\hline 
\end{tabular} 
\label{table1} 
\end{table}

\section{Proofs}\label{sec.proofs}

\no\textbf{Proof of Theorem~\ref{th.existenceU}.}
We divide the proof in several steps.

\noindent\textbf{Step 1. }Existence of a local in time solution to an auxiliary problem with smooth data. 

 We assume $u_0\in W^{1,\infty}(\O)$,  and 
 consider the following auxiliary problem, obtained using the change of unknown $u=w\text{e}^{\mu t}$ in \fer{eq.orig}, for some positive constant $\mu$ to be fixed:
\begin{align}
 \p_t w(t,\bx)  =& \int_\O \cK_h \big(\text{e}^{\mu t}(w(t,\by)-w(t,\bx))\big)(w(t,\by)-w(t,\bx)) d\by \nonumber \\ 
 & +\lambda(w_0(\bx)-w(t,\bx)) -\mu w(t,\bx), \label{eq.aux}
 \end{align}
 for $(t,\bx)\in (0,T_0)\times\O$, and 
for the initial data $w(0,\cdot)=w_0=u_0\in W^{1,\infty}(\O)$. Here, $T_0>0$ will be fixed
later.

\no\emph{Time discretization.} Let $N\in \N$, $\tau=T_0/N$ and $t_j=j\tau$, for $j=0,\ldots,N$. 
Assume that $w_j\in W^{1,\infty}(\O)$ is given and consider the functional $A:L^\infty(\O)\to L^\infty(\O)$ given by
\begin{align*}
A(\vfi(\bx))  = & \frac{1}{1+\tau(\lambda+\mu)} \Big(w_j(\bx) +\tau \int_{\O_*} \cK_h \big(\text{e}^{\mu t_j}(w_j(\by)-w_j(\bx))\big)(\vfi(\by)-\vfi(\bx)) d\by \\
    & +\tau \lambda w_0(\bx)\Big),
\end{align*}
 for $\bx \in \O$. Observe that if $A$ has a fixed point $\vfi$, then we may define $w_{j+1}=\vfi$ to get the following semi-implicit version of \fer{eq.aux}
 \begin{align}
 w_{j+1}(\bx)  = & w_{j}(\bx)+\tau \int_\O \cK_h \big(\text{e}^{\mu t_j}(w_{j}(\by)-w_{j}(\bx))\big)(w_{j+1}(\by)-w_{j+1}(\bx)) d\by   \nonumber \\
 & +\tau \lambda(w_0(\bx)-w_{j+1}(\bx)) -\tau \mu w_{j+1}(\bx).\label{eq.discrete}
 \end{align}
 We have, 
\begin{align*}
 \abs{A(\vfi(\bx)) -A(\psi(\bx))} & =\frac{\tau}{1+\tau\mu}
 \int_\O \cK_h \big(\text{e}^{\mu t_j}(w_{j}(\by)-w_{j}(\bx))\big) \\
  &\hspace{3.5cm} \times\big|\vfi(\by) -\psi(\by)-(\vfi(\bx) -\psi(\bx)) \big| d\by \\
 & \leq \frac{2\tau\abs{\O}}{1+\tau\mu}\nor{\cK_h}_\infty\nor{\vfi-\psi}_\infty.
\end{align*}
Therefore, for $\mu >2\abs{\O}\nor{\cK_h}_\infty$, the mapping $A$ is contractive in 
$L^\infty(\O_*)$, and a unique fixed point, $w_{j+1}$ verifying \fer{eq.discrete}
does exist. 

We have the following uniform estimates for $w_{j+1}$. 
One one hand, from \fer{eq.discrete} we obtain 
\begin{align*}
 \nor{w_{j+1}}_\infty \leq \frac{1}{1+\tau (\lambda +\mu -2\abs{\O}\nor{\cK_h}_\infty)}\Big(\nor{w_j}_\infty +\tau\lambda \nor{w_0}_\infty \Big), 
\end{align*}
which gives (recall $\mu > 2\abs{\O}\nor{\cK_h}_\infty$) the uniform estimate 
\begin{equation}
 \label{est.w}
 \nor{w_{j+1}}_\infty \leq M_0
\end{equation}
with $M_0$ depending only on $\nor{w_0}_\infty$.

On the other hand, since $w_0,w_j \in W^{1,\infty}(\O)$,  we deduce from \fer{eq.discrete}  $w_{j+1}\in W^{1,\infty}(\O)$. This regularity allows to 
differentiate  in \fer{eq.discrete} with respect to the $k-$th component of $\bx$, denoted by $x_k$, to obtain for a.e. $\bx\in\O$,
\begin{align*}
 F_1(\bx) \frac{\p w_{j+1}}{\p x_k}(\bx) = F_2(\bx) \frac{\p w_{j}}{\p x_k}(\bx) + \tau \lambda \frac{\p w_0}{\p x_k} (\bx), 
 \end{align*}
 with 
\begin{align*}
 F_1(\bx)  = & 1+ \tau \Big(\lambda+\mu  +\int_\O \cK_h \big(\text{e}^{\mu t_j}(w_{j}(\by)-w_{j}(\bx)) \big) d\by \Big),  \\
 F_2(\bx)  = & 1-\tau \text{e}^{\mu t_j}\int_\O \cK'_h \big(\text{e}^{\mu t_j}(w_{j}(\by)-w_{j}(\bx))\big)(w_{j+1}(\by)-w_{j+1}(\bx)) d\by , 
 \end{align*}
 from where we deduce
\begin{align*}
  \nor{\nabla w_{j+1}}_\infty \leq \frac{1}{1+ \tau (\lambda+\mu )} \Big( \tau \lambda \nor{\nabla w_ 0}_\infty  + 
  \big(1+ 2\tau\text{e}^{\mu t_j}\abs{\O}M_0 \nor{\cK'_h}_\infty  \big)  \nor{\nabla w_j}_\infty \Big). 
 \end{align*}
Solving this differences inequality, we find that, by redefining $\mu$ to satisfy $ \mu  > 2\text{e}^{\mu T_0}\abs{\O} \nor{\cK'_h}_\infty  M_0$,
we obtain the uniform estimate $\nor{\nabla w_{j+1}}_\infty\leq M_1$, with $M_1$ depending only on $\nor{\nabla w_0}_\infty$. This election of $\mu$ is possible by restricting $T_0$ to be 
\begin{equation}
 \label{def.T0}
 T_0<\frac{1}{\mu}\log\frac{\mu}{ 2\abs{\O} \nor{\cK'_h}_\infty  M_0}.
\end{equation}

\no\emph{Time interpolators and passing to the limit $\tau\to0$.} 
We define, for $(t,\bx)\in (t_j,t_{j+1}]\times \O$, the piecewise constant and
piecewise linear interpolators
\begin{align*}
 w^{(\tau)}(t,\bx)=w_{j+1}(\bx),\quad  
 \tilde w^{(\tau)}(t,\bx)=w_{j+1}(\bx)+\frac{t_{j+1}-t}{\tau}(w_j(\bx)-w_{j+1}(\bx)).
\end{align*}
Using the uniform $L^\infty$ estimates of $w_{j+1}$ and $\nabla w_{j+1}$, we deduce the corresponding uniform estimates for $\nor{ \nabla w^{(\tau)}}_{L^{\infty}(Q_{T_0})}$, 
$\nor{ \nabla \tilde w^{(\tau)}}_{L^{\infty}(Q_{T_0})}$ and
$\nor{ \p_t \tilde w^{(\tau)}}_{L^{\infty}(Q_{T_0})}$, implying the existence of 
$w\in L^\infty(0,{T_0}; W^{1,\infty}(\O))$ and 
$\tilde w\in W^{1,\infty}(Q_{T_0})$ such that, at least in a subsequence (not relabeled), as $\tau \to 0$,
\begin{align}
 & w^{(\tau)} \to w \qtext{weakly* in }L^\infty(0,T_0; W^{1,\infty}(\O)), \nonumber \\
 & \tilde w^{(\tau)} \to \tilde w \qtext{weakly* in }   W^{1,\infty}(Q_{T_0}). \label{conv.1}
\end{align}
In particular, by compactness 
\begin{equation*}
 \tilde w^{(\tau)} \to \tilde w \qtext{uniformly in } C([0,T_0]\times \bar \O). 
\end{equation*}
Since, for $t\in (t_j,t_{j+1}]$,
\begin{align*}
 \abs{ w^{(\tau)}(t,\bx) -\tilde w^{(\tau)}(t,\bx) }= \abs{\frac{(j+1)\tau -t}{\tau}(w_j(\bx)-w_{j+1}(\bx))}\leq \tau \nor{\p_t \tilde w^{(\tau)}}_{L^{\infty}(Q_{T_0})},
\end{align*}
we deduce both $w=\tilde w$ and 
\begin{equation}
 \label{conv.2}
 w^{(\tau)} \to  w \qtext{uniformly in } C([0,T_0]\times \bar \O). 
\end{equation}
Considering the shift operator $\sigma_\tau w^{(\tau)}(t,\cdot) = w_{j}$, 
and introducing the approximation $\text{e}_\tau^{\mu t}=\text{e}^{\mu t_j}$,
for  $t\in(t_j,t_{j+1}]$, we may rewrite \fer{eq.discrete} as 
\begin{align}
 \p_t \tilde w^{(\tau)}(t,\bx)  & = \int_\O \cK_h \big(\text{e}_\tau^{\mu t}( w^{(\tau)}(t,\by)- w^{(\tau)}(t,\bx))\big)(w^{(\tau)}(t,\by)-w^{(\tau)}(t,\bx)) d\by  \nonumber \\
 & + \lambda(w_0(\bx)-w^{(\tau)}(t,\bx)) - \mu w^{(\tau)}(t,\bx), \label{eq.final}
 \end{align}
 and due to the convergence properties \fer{conv.1} and \fer{conv.2}, we may pass to the limit $\tau\to0$ in \fer{eq.final} to deduce that $w$ is a solution of 
 \fer{eq.aux}.
 
  \noindent\emph{Continuation of the solution to an arbitrary time $T$.} 
 Given the solution, $w$, of problem \fer{eq.aux} in $Q_{T_0}$, we may consider the same problem 
 for the initial datum $w(T_0,\cdot)$. Since $w(T_0,\cdot)\in W^{1,\infty}(\O)$ and the constant
 $T_0>0$ only depends on $\abs{\O}$, $\nor{\cK'_h}_\infty$ and $\nor{u_0}_\infty$, 
 see \fer{est.w} and \fer{def.T0}, we obtain a new solution 
  $w\in C(T_0,2T_0;W^{1,\infty}(\O))$. Clearly, this procedure may be 
  extended to an arbitrarily fixed $T$. Once this is done, a boot-strap argument allows us to deduce $w\in C^\infty(0,T;W^{1,\infty}(\O))$, implying that $u=w\text{e}^{\mu t} \in C^\infty([0,T];W^{1,\infty}(\O))$
 is a solution of \P~in $Q_{T}$.

\noindent\textbf{Step 2. }Non smooth initial data. 

Let us consider a sequence 
$u_{0\eps}\in C^\infty(\bar \O)$
such that, as $\eps\to0$, 
\begin{align}
 & u_{0\eps} \to u_0 \qtext{in }L^\infty(\O) \label{reg1.id} \\
& \nor{\nabla u_{0\eps}}_{L^1(\O)}\to \text{TV}(u_0), \label{reg2.id}
\end{align}
where TV denotes total variation with respect to the $\bx$ variable. Let us denote by  $u_\eps$ to the corresponding solution of P$(\O,u_{0\eps})$. 

First, notice that $u_\eps$ is uniformly bounded in $L^\infty(Q_T)$ with respect to $\eps$
as a consequence of estimate \fer{est.w} and property \fer{reg1.id}. We then obtain directly from equation \fer{eq.orig} that 
\begin{equation}
 \label{bound.1}
\partial_t u_\eps \qtext{is uniformly bounded in }L^\infty(Q_T). 
\end{equation}
Since $u_{0\eps}$ is smooth, we may deduce an $L^\infty$ bound for $\grad u_\eps$ as in Step 1, not necessarily uniform in $\eps$, but which allows us to differentiate equation \fer{eq.orig} with respect to $x_k$. After integration in $(0,t)$,  we obtain
  \begin{align}
 \label{dsv}
  \frac{\p u_\eps}{\p x_k} (t,\bx)= \frac{\p u_{0\eps}}{\p x_k}(\bx) G_\eps(t,\bx) \Big(  1 + 
  \lambda \int_0^t (G_\eps(\tau,\bx))^{-1}  d\tau \Big),
 \end{align}
 with $G_\eps(t,\bx)=  \exp\big( -\int_0^t \big( \lambda + \eta_\eps(\tau,\bx)\big)  d\tau \big)$, and 
 \begin{align}
 \label{eta}
  \eta_\eps(t,\bx) =& \int_{\O} \Big( \cK'_h \big( u_\eps(t,\by)-u_\eps(t,\bx)\big) ( u_\eps(t,\by)-u_\eps(t,\bx)) \\
  & + \cK_h \big( u_\eps(t,\by)-u_\eps(t,\bx)\big) \Big)d\by .\nonumber
 \end{align}
Since $\cK_h\in W^{1,\infty}(\R)$, we have 
$\eta_\eps$ uniformly bounded in $L^\infty(Q_T)$ and so $G_\eps$ and $G_\eps^{-1}$. Therefore, using \fer{reg2.id} 
  we deduce from \fer{dsv} that 
\begin{equation}
 \label{bound.2}
\nabla u_\eps \qtext{is uniformly bounded in }L^\infty(0,T;L^1(\O)). 
\end{equation}
Bounds \fer{bound.1} and \fer{bound.2} allow to deduce, using the compactness result \cite[Cor. 4, p. 85]{Simon1986},  the existence of $u\in C([0,T];\X)$
  such that $u_\eps\to u$ strongly in  $L^p(Q_T)$, for all $p<\infty$, and a.e. in $Q_T$. 
  
  Similarly to the smooth case, this convergence allows to pass to the limit $\eps\to 0$ in \fer{eq.orig} (with $u$ replaced by $u_\eps$) and identify the limit $u$ as a solution of \P. Again, the property  $u\in L^\infty(Q_T)$ and a boot-strap argument leads to $u\in C^\infty (0,T;\X)$.
  
  \noindent\emph{Stability and uniqueness. }
  Let $u_{01},~u_{02}\in BV(\O)$ and $u_1,~u_2 \in C^\infty([0,T];\X)$ be the corresponding solutions to problems $P(\O,u_{01}),~P(\O,u_{02})$. Set $u=u_1-u_2$ and
  $u_0=u_{10}-u_{20}$. Then $u$ satisfies  
  \begin{align*}
   \p_t u(t,\bx)  = &
 \int_\O \Big( \Phi(u_1(t,\by)-u_1(t,\bx))- \Phi(u_2(t,\by)-u_2(t,\bx))\Big)d\by \\
   & +\lambda (u_0(\bx)-u(t,\bx)), \\
    u(0,\bx)=& u_0(\bx),
 \end{align*}
 for $(t,\bx)\in Q_T$, with $\Phi(s)=\cK_h(s)s$. Multiplying this equation by $u$, integrating in $\O$ and using the Lipschitz continuity of $\Phi$ (with constant $C_L$) and Young's inequality, we deduce
  \begin{align*}
  \p_t \int_\O \abs{u(t,\bx)}^2d\bx   \leq & C_L 
 \int_\O\int_\O  \abs{u(t,\by)-u(t,\bx)}u(t,\bx) d\by d\bx +\frac{\lambda}{2}\int_\O \abs{u_0(\bx)}^2 d\bx  \\ 
 & -\frac{\lambda}{2}\int_\O \abs{u(t,\bx)}^2 d\bx  \\
 & \leq  C_L \Big(\int_\O \abs{u(t,\bx)}d\bx \Big)^2  +(C_L-\frac{\lambda}{2}) \int_\O  \abs{u(t,\bx)}^2 d\bx \\
    & +\frac{\lambda}{2}\int_\O \abs{u_0(\bx)}^2 d\bx.
 \end{align*}
 Finally, using Jensen's and Gronwall's inequalities, we deduce $\nor{u(t,\cdot)}_{L^2(\O)}\leq C \nor{u_0}_{L^2(\O)}$ for all $t\in (0,T)$, and the result follows.
 
  \noindent\emph{Time invariance of the level sets. }The proof of this property is similar to the proof of the stability property. 
  Let $u_{0}\in \X$ and $u \in C^\infty([0,T];\X)$ be the corresponding solution to problem $P(\O,u_{0})$. Assume $u_0(\bx_1)=u_0(\bx_2)$, and set $u_i(t)= u(t,\bx_i)$, $i=1,2$. 
 Then, from equation \fer{eq.orig} we get 
  \begin{align*}
   \p_t (u_1(t)-u_2(t))  = &
 \int_\O \Big( \Phi(u(t,\by)-u_1(t))- \Phi(u(t,\by)-u_2(t))\Big)d\by \\
    & -\lambda (u_1(t))-u_2(t)).
  \end{align*}
Then, the Lipschitz continuity of $\Phi$ and  Gronwall's lemma allow us to deduce the result.
$\Box$
  
  \bigskip
  
  \no\textbf{Proof of Corollary~\ref{th.existenceV}.}  
 To prove point 1, notice that from  \fer{dsv} (in dimension $d=1$) we deduce
  \begin{equation*}
   \sign(\p_s v_\eps) = \sign(v'_{0\eps})\leq 0 \qtext{a.e. in }(0,T)\times\O_*,
  \end{equation*}
a property that also holds in the limit $\eps\to 0$. Point 2 of the theorem 
follows from evaluating equation \fer{eq.orig} in $s=0$ and $s=\abs{\O}$, using that 
$v(t,\cdot)$ is decreasing for all $t>0$, and Gronwall's inequality. Point 3 is a consequence of the assumption on the symmetry of $\cK$, under which the integral term in \fer{eq.orig} vanishes when it is integrated in $\O_*$. Point 4 is easily deduced by successive derivation of \fer{dsv} (which also holds for $\eps=0$, under regularity assumptions). Point 5 is again deduced from \fer{dsv} and the decreasing character of $v_\eps$ and $v$. Since, $TV(v_\eps(t,\cdot))\to TV(v(t,\cdot))$ and, 
using point 2, $TV(v(t,\cdot))\leq c$ for all $t\geq 0$, we have that the integral term in \fer{eta} is evaluated inside a closed interval. Therefore, using the assumptions of point 5, we get $\eta_\eps (t,s) >c_2>0$ uniformly in $(t,s)$. Finally, we obtain the result from \fer{dsv} in the limit $t\to\infty$ and $\eps\to 0$.  
$\Box$

\bigskip  
  
\no\textbf{Proof of Theorem~\ref{th.equiv}.}
We split the proof in two steps.

\no\textbf{Step 1.} First we treat the case in which $u_0$ has no flat regions, that is when $\abs{\{\bx \in\O : u(\bx) =q\}} =0$ for any $q\in\R$. By the invariance of the level sets structure proven in Theorem~\ref{th.existenceU} we deduce that neither the solution $u$ of \P~has  flat regions. Then $m_u(t,\cdot)$ and $u_*(t,\cdot)$ are strictly decreasing, implying $u_*(t,m_u(t,q))=q$ for any $q\in\R$.   According to  \cite[Theorem~9.2.1]{Rakotoson2008}, we have $\p_t u_*=\p_s \vfi$
where 
\begin{equation}
\label{cor.1}
 \vfi(t,s) =\int_{\{u(t)>u_*(t,s)\}} \frac{\p u}{\p t}(t,\bx)d\bx,
\end{equation}
and we used the notation $\{u(t)>u_*(t,s)\} = \{\by\in\O: u(t,\by)>u_*(t,s)\}$.
Integrating \fer{eq.orig} in $\{u(t)>u_*(t,s)\}$ we get
\begin{align}
\label{cor.2}
 \vfi(t,s)  = & \int_{\{u(t)>u_*(t,s)\}}
 \int_\O \cK_h (u(t,\by)-u(t,\bx))(u(t,\by)-u(t,\bx)) d\by d\bx \\
   & +\lambda \int_{\{u(t)>u_*(t,s)\}} u_0(\bx)d\bx -\lambda \int_{\{u(t)>u_*(t,s)\}}u(t,\bx)d\bx =I_1+I_2+I_3. \nonumber
\end{align}
Due to the $u$ and $u_*$ level sets equi-measure, it is immediate that 
\begin{align}
\label{cor.3}
I_3=-\lambda \int_0^s u_*(t,\sigma)d\sigma.
\end{align}
The equi-measurability property \fer{prop.1}  implies
\begin{align*}
I_1  =   \int_{\{u(t)>u_*(t,s)\}} \int_{\O_*}
  \cK_h (u_*(t,\sigma)-u(t,\bx))(u_*(t,\sigma)-u(t,\bx)) d\sigma d\bx,
\end{align*}
from where we deduce
\begin{align}
 \label{cor.4}
I_1  = \int_0^s \int_{\O_*}  
\cK_h (u_*(t,\sigma)-u_*(t,\tau))(u_*(t,\sigma)-u_*(t,\tau)) d\sigma d\tau.
\end{align}
To deal with the term $I_2$ we observe that due to the invariance of the level set structure, as stated in Theorem~\ref{th.existenceU}, we have that, for all $t\in [0,T]$
and $s\in\bar\O_*$, there exists $\alpha\in\bar\O_*$ such that
\begin{equation*}
 \{\bx\in\O: u(t,\bx)>u_*(t,s)\}=\{\bx\in\O:u_0(\bx)>u_{0*}(\alpha)\}.
\end{equation*}
Recalling that $u$ and $u_0$ have not flat regions and taking the measure 
of these sets we deduce $s=\alpha$. Therefore,
\begin{align}
\label{cor.5}
 I_2=\lambda \int_{\{u_0>u_{0*}(s)\}} u_0(\bx)d\bx = \int_0^s u_{0*}(\sigma)d\sigma.
\end{align}
Finally, substituting in identity \fer{cor.2} the expressions \fer{cor.1}, \fer{cor.3}, 
\fer{cor.4} and \fer{cor.5}, and differentiating with respect to $s$, we deduce the result.
  
\bigskip

Conversely, let $v$ be a solution of \Ps. 
Since $u_0$ has not flat regions, $u'_{0*}<0$ in $\O_*$, and by point 1 
of Corollary~\ref{th.existenceV} we have $\p_s v(t,s) <0$ in $[0,T]\times \O_*$.
We define 
\begin{equation}
\label{def.u}
 u(t,\bx)=v(t,s)\qtext{for a.e. }\bx\in L(s)=\{ \by\in\O: u_0(\by)=u_{0*}(s)\},
\end{equation}
and for all $t\in [0,T]$. Observe that since $u_0$ has not flat regions, we have
$\abs{L(s)}=0$ for all $s$. Therefore, since $\p_s v <0$, we also deduce that $u$ has not flat regions.
By construction, 
\begin{equation*}
 \left| \{ \bx\in\O: u(t,\bx)>v(t,s) \}\right| = 
 \left| \{ \bx\in\O: u_0(\bx)>u_{0*}(s) \}\right| =s,
\end{equation*}
implying $u_*=v$. Differentiating in \fer{def.u} with respect to $t$ and using that $v$
is a solution of \Ps, we get, for $\bx\in L(s)$,
\begin{align*}
 \p_t u (t,\bx)& = \p_t v(t,s)  =\int_{\O_*} \cK_h (v(t,\sigma)-v(t,s))(v(t,\sigma)-v(t,s)) d\sigma   +\lambda(v_0(s)-v(t,s)) \\
 & = \int_{\O_*} \cK_h (u_*(t,\sigma)-u (t,\bx))(u_*(t,\sigma)-u (t,\bx)) d\sigma   +\lambda(u_0(\bx)-u (t,\bx))\\
 & = \int_\O \cK_h (u(t,\by)-u (t,\bx))(u(t,\by)-u (t,\bx)) d\by   +\lambda(u_0(\bx)-u (t,\bx)),
\end{align*}
where we have used again the equi-measurability property \fer{prop.1}.  

\no\textbf{Step 2.} We now treat the general case in which $u_0 \in \X$  may have flat regions. We use the following lemma.
\begin{lemma}
\label{lemma.approx}
 Let $u_0\in \X$. Then there exists a sequence $u_{0j}\in \X$ such that $u_{0j}$ has no flat regions and $u_{0j}\to u_0$ in $\X$.
\end{lemma}

We may then apply the Step 1 of this proof to each $u_{0j}$ to obtain that 
$u_j \in C^\infty([0,T];\X)$ is a solution of $P(\O,u_{0j})$ (without flat regions)
if and only if $(u_j)_* \in C^\infty(0,T;\X_*)$ is a  solution of $P(\O_*,(u_{0j})_*)$. Now we perform the limit $j\to\infty$.

Let $u \in C^\infty([0,T];\X)$ and $v \in C^\infty(0,T;\X_*)$ be the solutions of
problems $P(\O,u_{0})$ and $P(\O_*,u_{0*})$  ensured by 
Theorem~\ref{th.existenceU}. 

Using the strong continuity of the decreasing rearrangement operation in 
$L^2(\O)$, see \fer{prop.2}, and the stability property \fer{stability} applied to problem $P(\O,u_{0})$, we obtain
\begin{equation*}
 \nor{u_*-(u_j)_*}_{L^\infty(0,T;L^2(\O_*))} \leq \nor{u-u_j}_{L^\infty(0,T;L^2(\O))}\leq C_1\nor{u_{0}-u_{0j}}_{L^2(\O)}.
\end{equation*}
The same arguments in reverse order  applied to problem $P(\O_*,u_{0*})$ leads to 
\begin{equation*}
 \nor{v-(u_{j})_*}_{L^\infty(0,T;L^2(\O_*))}\leq C_2 \nor{u_{0*}-(u_{0j})_*}_{L^2(\O_*)}\leq C_2\nor{u_{0}-u_{0j}}_{L^2(\O)}.
\end{equation*}
Therefore, using the triangle inequality we deduce
\begin{align*}
 \nor{v-u_*}_{L^\infty(0,T;L^2(\O_*))}  \leq (C_1+C_2)\nor{u_{0}-u_{0j}}_{L^2(\O)} \to 0,
\end{align*}
as $j\to\infty$. $\Box$

\bigskip

\no\textbf{Proof of Lemma~\ref{lemma.approx}. }

In this proof, we rename $u_{0}$ by $u$ and $u_{0j}$ by $u_j$. 
Let, for $i\in I$, $E_i=\{\bx\in\O: u(\bx)=q_i\}$ with $\abs{E_i}>0$,  be the collection of flat regions of $u$ which is, at most, countable. Thus, $I\subset\N$. Let
$\Chi{E_i}$ and $P(E_i)$ denote the characteristic function of the set $E_i$ and its perimeter, respectively. 
We consider the functions
\begin{equation*}
 \vfi_i^j(\bx)=\frac{\Chi{E_i}(\bx)\min(q_i-q_{i+1},1)}{i^2\big(j(1+P(E_i))+v(\bx)\big)},
\end{equation*}
where
$v\in BV(\O)$ is a non-negative function without flat regions. 
Observe that since $u\in BV(\O)$ we have $P(E_i)<\infty$ for all $i\in I$.

 Consider, for $j\in \N$, the sequence of $L^\infty(\O)$ functions 
\begin{equation*}
 u_{j}(\bx)=u(\bx) -\sum_{i\in I} \vfi_i^j(\bx)=\left\{
 \begin{array}{ll}
  u(\bx)&\text{if }\bx \in \O\backslash \cup_{i\in I} E_i\\
  q_i - \frac{\min(q_i-q_{i+1},1)}{i^2\big(j(1+P(E_i))+v(\bx)\big)} & \text{if }\bx\in E_i, \text{for some }i\in I.
 \end{array}
 \right.
\end{equation*}
We have:

\no(1) \emph{$u_{j}$ has no flat regions in $\O$}. Let $q\in\R$. We use the decomposition
\begin{equation*}
 \{\bx\in\O: u_{j}(\bx)=q\}=\{\bx\in \O\backslash \cup_i E_i: u(\bx)=q\} \bigcup \cup_i 
 \{\bx\in E_i: u_{j}(\bx)=q\}.
\end{equation*}
If $q=q_i$ for some $i\in I$ then $\bx\in E_i$, and, by definition, $u_{j}(\bx)=q_i$ if
\[
 \frac{1}{i^2\big(j(1+P(E_i))+v(\bx)\big)}=0,
\]
which is not possible. Therefore, if $q=q_i$ we have $\abs{u_{j}=q_i}=0$.
If $q\neq q_i$ for all $i\in I$ then  $\abs{\{\bx\in \O\backslash \cup_i E_i: u(\bx)=q\}}=0$, so 
\[
\abs{u_{j}(\bx)=q}=\abs{ v(\bx)= -j(1+P(E_i))+\min(q_i-q_{i+1},1)/(i^2(q_i-q))}=0,
\]
since $v$ has no flat regions.

\no(2) \emph{$u_{j}\to u$ in $L^p(\O)$ for any $1\leq p\leq \infty$}. This is immediate, since 
$\abs{u(\bx)-u_{j}(\bx)}\leq \frac{1}{j}$.

\no(3) \emph{$u_{j}\in BV(\O)$  and $u_{j}\to u$ in $BV(\O)$.} 
According to \cite[Proposition 3.38]{Ambrosio2000}, for each $i\in I$, we can find a sequence $w_h^i\in C^\infty(\O)$ with $0\leq w_h^i\leq 1$ such that $w_h^i\to \Chi{E_i} $ in $L^1(\O)$ as $h\to0$, and 
\begin{equation*}
 \lim_{h\to 0} \int_\O \abs{\grad w_h^i}  = TV(\Chi{E_i})=P(E_i)<\infty,
\end{equation*}
since $u\in BV(\O)$. We also introduce a regularizing sequence $v_h\in C^\infty(\O)$ such that $v_h>0$ and $v_h\to v$ in $BV(\O)$. Let $g_h^i= w_h^i \min(q_i-q_{i+1},1)/(i^2(j(1+P(E_i))+v_h))$. Then,
\begin{align}
\label{est.bv}
 \int_\O \abs{\grad g_h^i}\leq \frac{1}{i^2j(1+P(E_i))} \int_\O \abs{\grad w_h^i} + \frac{1}{(ij(1+P(E_i)))^2}\int_\O \abs{\grad v_h} ,
\end{align}
implying that $g_h^i$ is uniformly bounded in $BV(\O)$ with respect to $h$. Therefore, there exists $g^i\in BV(\O)$ and a  a subsequence of $g_h^i$ (not relabeled) such that $g_h^i\to g^i$ strongly in $L^1(\O)$ as $h\to0$. 
Since, by the Dominated Convergence Theorem we have  $g_h^i\to \vfi_i^j$ in $L^1(\O)$ as $h\to0$, we deduce $g^i=\vfi_i^j\in BV(\O)$. Taking the limit $h\to0$ in  \fer{est.bv} we get
\begin{equation}
\label{lemma.1}
 TV(\vfi_i^j)\leq  \frac{P(E_i)}{i^2j(1+P(E_i))} + \frac{1}{(ij(1+P(E_i)))^2}\int_\O \abs{\grad v_h} \leq \frac{c}{i^2j}, 
\end{equation}
with $c>0$ independent of $i$ and $j$. Thus, using the definition of $u_j$, the triangle inequality, and \fer{lemma.1} we get
\begin{equation*}
 TV(u_{j})\leq TV(u)+ \sum_{i\in I} TV(\vfi_i^j) \leq TV(u)+\frac{c}{j}.
\end{equation*}
Therefore, $u_j\in BV(\O)$. Finally, from the definition of $u_j$ and \fer{lemma.1} we have 
\begin{equation*}
 TV(u-u_{j}) \leq \sum_{i\in I} TV(\vfi_i^j) \to 0 \qtext{as }j\to \infty.
\end{equation*}
$\Box$

 \bigskip

\no\textbf{Proof of Theorem~\ref{th.pde}. }
Define
\begin{align}
\label{eq.i}
 I(t,s)  =\int_{\O_*} \cK_h (u_*(t,\sigma)-u_*(t,s))(u_*(t,\sigma)-u_*(t,s)) d\sigma.
 \end{align}
Since $u_0$ has not flat regions, we have $u_{0*}' <0$. Then, due to points 1 and 2 of Corollary~\ref{th.existenceV} we have $\p_s u_*<0$ in $[0,T)\times \O_*$, and $u_*(t,\O_*)\subset u_{0*}(\O_*)$ for all $t\in [0,T]$, respectively.

Let us consider the inverse of $u_*(t,\cdot)$, the distribution function of $u$, $m_{u}(t,\cdot)$.
Using the change of variable $s=m_{u}(\cdot,z)$ and writing
$\sigma=m_{u}(\cdot, q)$, we obtain from \fer{eq.i}
\begin{equation}
\label{th2.3}
I_1(t,z):=I(t,m_{u}(t,z))=
\int_{u_*(t,\abs{\O})}^{u_*(t,0)}  \cK_h(q-z)(q-z)\frac{dq}{\abs{\p_s u_* (t, m_{u}(t,q))}}.
\end{equation}
Using the explicit form of $\cK$ and integrating by parts, we obtain
\begin{equation}
\label{th2.i1}
 I_1(t,z)=\frac{h^2}{2}\Big( \tilde k_h (m_{u}(t,z))
 +\int_{u_*(t,\abs{\O})}^{u_*(t,0)} \cK_h(q-z)\frac{\p^2_{ss}u_* (t,m_{u}(t,q))}{(\p_su_*(t,m_{u}(t,q)))^3}dq\Big),
\end{equation}
with $\tilde k_h$ given by \fer{def.ktilde}.

By assumption, function
\begin{equation*}
 f(t,q)=\frac{\p^2_{ss}u_*(t,m_{u}(t,q))}{(\p_s u_*(t,m_{u}(t,q)))^3} 
\end{equation*}
is bounded in $[u_* (t, (\abs{\O})),u_*(t, 0)]$ and by point 4 of Corollary~\ref{th.existenceV} it is continuously differentiable in $(u_*(t, \abs{\O}),u_*(t, 0))$.

Consider the interval $J_h=\{q: \abs{q-z}<\sqrt{h}\}$.
By well known properties of the Gaussian kernel, we have
\begin{equation}
 \label{gauss.1}
 \kappa(h): = \int_{J_h} \cK_h(q-z) dq <  \int_\R \cK_h(q) dq = h\sqrt{\pi},
\end{equation}
and
\begin{equation}
 \label{gauss.2}
\cK_h(z-q)\leq  \text{e}^{-1/h} \quad\text{if}\quad  q\in J_h^C=\{q:\abs{q-z}\geq\sqrt{h}\}.
\end{equation}
 In particular, from \fer{gauss.2} we get 
\begin{equation}
\label{th2.4}
 \left| \int_{J_h^C} \cK_h(q-z)f(t,q)dq \right| < O(h^\alpha)\qtext{for any }\alpha >0.
\end{equation}
Taylor's formula implies 
\begin{align*}
 \int_{u_*(\abs{\O})}^{u_*(0)}  \cK_h(q-z)f(t,q)dq & =
 \int_{J_h} \cK_h(q-z) (f(t,z)+O(\sqrt{h})) dq \\
 & + \int_{J_h^C} \cK_h(q-z)f(t,q)dq .
\end{align*}
Therefore, from \fer{th2.i1}, \fer{th2.4} and \fer{gauss.1} we deduce, using $\p_s u_*<0$,
\begin{align*}
I_1(t,z)=\frac{h^2}{2} \Big( \tilde k_h(m_{u}(t,z)) -
\frac{\p^2_{ss} u_* (t,m_{u}(t,z))}{\abs{\p_s u_* (t,m_{u}(t,z))}^3} \kappa(h) + O(h^{3/2})\Big).
\end{align*}

Then, the result follows from \fer{th2.3} substituting $z$ by $u_*(t,s)$. $\Box$

\section{Conclusions}\label{sec.conclusions}
In this paper we studied a general class of nonlinear integro-differential  operators with important  imaging applications,  such as the denoising-segmentation Neighborhood filtering.

Although the corresponding PDE problem is multi-dimensional, we showed that it can be 
reformulated as a one-dimensional problem by means of the notion and properties of the decreasing rearrangement function.
We proved the well-posedness of the problem and some stability properties of the solution,  as well as the 
equivalence between the multi-dimensional and the one-dimensional solutions to the problem.

Some other interesting properties were deduced for the rearranged one-dimensional version of the problem, such as the time invariance of the level sets of the solution (inherited by the multi-dimensional equivalent solution), and the asymptotic behavior of the solution as a shock-type filter.

Future work will point to the use of rearranging techniques for the generalization of the model to include nonlocal effects induced by non-homogeneous spatial kernels, like in equation 
\fer{mazon}. As already showed for the discrete time problem \cite{Galiano2015b}, this situation is much more involved suggesting  the consideration of the {\em relative rearrangement} functional.

\end{document}